\documentclass{article}

\usepackage{arxiv}

\usepackage[utf8]{inputenc}
\usepackage[T1]{fontenc}
\usepackage{amsmath,amssymb,amsfonts}
\usepackage{float}
\usepackage{caption}
\usepackage{subcaption}

\usepackage{booktabs}
\usepackage{makecell}
\usepackage{multirow}
\usepackage{siunitx}

\usepackage{pifont}
\newcommand{\xmark}{\ding{55}}
\newcommand{\cmark}{\ding{51}}

\usepackage{authblk}
\usepackage{orcidlink}

\usepackage{url}
\usepackage{hyperref}
\hypersetup{
    colorlinks=true,
    linkcolor=blue,
    citecolor=blue,
    urlcolor=blue,
    pdfborder={0 0 0}
}

\usepackage{cite}
\usepackage{cleveref}

\usepackage{glossaries}
\glsdisablehyper
\newacronym{ai}{AI}{Artificial Intelligence}
\newacronym{cti}{CTI}{Cyber Threat Intelligence}
\newacronym{llm}{LLM}{Large Language Model}
\newacronym{misp}{MISP}{Malware Information Sharing Platform}
\newacronym{shacl}{SHACL}{Shapes Constraint Language}
\newacronym{stix}{STIX}{Structured Threat Information Expression}
\newacronym{ttp}{TTP}{Tactics, Techniques, and Procedures}
\newacronym{uco}{UCO}{Unified Cyber Ontology}
\newacronym{kg}{KG}{Knowledge Graph}
\newacronym{rag}{RAG}{Retrieval Augmented Generation}
\newacronym{rq}{RQ}{Research Questions}
\newacronym{mmr}{MMR}{Maximal Marginal Relevance}
\newacronym{rdf}{RDF}{Resource Description Framework}

\usepackage{listings}
\usepackage[frozencache,cachedir=.]{minted}
\newcommand{\footurl}[1]{\footnote{\url{#1} (on January 25, 2025).}}
\newcommand{\mitreattack}[0]{MITRE ATT\&CK}
\newcommand{\llm}[1]{#1}
\newcommand{\claudesonnet}[0]{\llm{Claude Sonnet~4}}
\newcommand{\claudehaiku}[0]{\llm{Claude~3.5 Haiku}}
\newcommand{\llamasmall}[0]{\llm{Llama~3.1 (8B)}}
\newcommand{\llamabig}[0]{\llm{Llama~3.3 (80B)}}
\newcommand{\mistrallarge}[0]{\llm{Mistral Large (123B)}}
\newcommand{\gptosssmall}[0]{\llm{gpt-oss (20B)}}
\newcommand{\gptossbig}[0]{\llm{gpt-oss (120B)}}
\newcommand{\qwencoder}[0]{\llm{Qwen3 Coder (32B)}}

\title{OntoLogX:\@ Ontology-Guided Knowledge Graph Extraction from Cybersecurity Logs with Large Language Models}

\author[1,*]{Luca Cotti\,\orcidlink{0009\--0004\--6351\--556X}\,}
\author[2]{Idilio Drago\,\orcidlink{0000\--0003\--1932\--1261}\,}
\author[1]{Anisa Rula\,\orcidlink{0000\--0002\--8046\--7502}\,}
\author[1]{Devis Bianchini\,\orcidlink{0000\--0002\--7709\--3706}\,}
\author[1,3,4]{Federico Cerutti\,\orcidlink{0000\--0003\--0755\--0358}\,}
\affil[1]{Department of Information Engineering, University of Brescia, Italy}
\affil[2]{Department of Computer Science, University of Turin, Italy}
\affil[3]{School of Computer Science and Informatics, Cardiff University, United Kingdom}
\affil[4]{Department of Electronics and Computer Science, University of Southampton, United Kingdom}
\affil[*]{Corresponding author. Email: \href{mailto:luca.cotti@unibs.it}{luca.cotti@unibs.it}}

\begin{document}

\maketitle

\begin{abstract}
    System logs represent a valuable source of \gls{cti}, capturing attacker behaviors, exploited vulnerabilities, and traces of malicious activity. Yet their utility is often limited by a lack of structure, semantic inconsistency, and fragmentation across devices and sessions. Extracting actionable \gls{cti} from logs, therefore, requires approaches that can reconcile noisy, heterogeneous data into coherent and interoperable representations. We introduce \textit{OntoLogX}, an autonomous \glsunset{ai}\gls{ai} agent that leverages \glspl{llm} to transform raw logs into ontology-grounded \glspl{kg}. OntoLogX integrates a lightweight log ontology with \gls{rag} and iterative correction steps, ensuring that generated \glspl{kg} are syntactically and semantically valid. Beyond event-level analysis, the system aggregates \glspl{kg} into sessions and employs a \gls{llm} to predict \mitreattack\ tactics, linking low-level log evidence to higher-level adversarial objectives. We evaluate OntoLogX on both public and real-world honeypot datasets, demonstrating robust \gls{kg} generation across multiple \glspl{llm} backends and accurate mapping of adversarial activity to \mitreattack\ tactics. Results highlight the effectiveness of the methodology in constructing ontology-compliant \glspl{kg}, along with their value in extracting actionable \gls{cti}.
\end{abstract}


\section{Introduction}\label{sec:introduction}
The rapid evolution and increasing sophistication of cyber threats pose significant risks to individuals, organizations, and governments, as adversaries continuously adapt their tactics to exploit vulnerabilities and avoid detection~\cite{liComprehensiveReviewStudy2021, thakurInvestigationCyberSecurity2015,scalaRiskFiveHard2019}. Traditional reactive defenses, such as signature- or rule-based systems, often fail to keep pace with this dynamic landscape, motivating a shift toward proactive and anticipatory strategies~\cite{hutchinsIntelligencedrivenComputerNetwork2011, papernotSoKSecurityPrivacy2018}.

\gls{cti}, defined as the collection, processing, and analysis of information about threat actors' motives, targets, and attack behaviors, supports faster and better-informed decisions in cybersecurity operations. By enabling a transition from reactive to proactive defense, \gls{cti} strengthens the ability of organizations, governments, and individuals to anticipate and mitigate attacks~\cite{tounsiSurveyTechnicalThreat2018,sunCyberThreatIntelligence2023}. Among the various sources of \gls{cti}, system logs are especially valuable, and especially those produced by honeypot and honeynet deployments. Unlike operational logs dominated by benign activity, honeypots are designed to attract and record malicious interactions, resulting in data with a higher concentration of adversarial behavior~\cite{nawrockiSurveyHoneypotSoftware2016}.

Processing logs, however, is a difficult task, as they are typically unstructured, syntactically heterogeneous, and often ambiguous in meaning, which complicates automated analysis. Moreover, the information required to reconstruct attack scenarios is frequently fragmented across multiple logs, which may be distributed over different devices. Traditional rule-based or heuristic techniques lack the adaptability to generalize across diverse and evolving threat behaviors.

To address these challenges, recent works have explored the construction of \glspl{kg} as a means to structure and enrich \gls{cti}~\cite{zhangAttacKGBoostingAttack2025,huangMITREtrievalRetrievingMITRE2024,falcarinBuildingCybersecurityKnowledge2024,liuConstructingKnowledgeGraph2023}.\ \glspl{kg} describe the concepts, entities, and events of the objective world, as well as the relations between them, and express knowledge in a form closer to that of human cognition, compared to other organizational representations~\cite{zhaoSurveyCybersecurityKnowledge2024}. Such representations facilitate semantic reasoning, integration with automated workflows, and support for advanced analytical tasks.

At the same time, advances in \glspl{llm} have shown remarkable effectiveness in extracting structured information from natural language~\cite{izacardLeveragingPassageRetrieval2021,lewisRetrievalaugmentedGenerationKnowledgeintensive2020,srivastavaImitationGameQuantifying2023}, fueling widespread adoption across diverse domains~\cite{zhangWhenLLMsMeet2025}. Yet, applications of \glspl{llm} in \gls{cti} remain limited, with existing approaches relying on heavily pre-processed inputs or substantial user interaction~\cite{zhangAttacKGBoostingAttack2025,payneLogFileAnomaly2024}.

In this work, we present \textit{OntoLogX}, an autonomous \gls{ai} agent designed to extract \gls{cti} from raw logs. OntoLogX leverages an \gls{llm} to construct detailed \glspl{kg} that capture both the structural and contextual aspects of the raw log events, without requiring user intervention. The generated graphs conform to a domain-specific ontology tailored for cybersecurity logs, enabling semantic querying, traceability, and automated reasoning.\ This standardized representation facilitates further analysis and integration with existing CTI systems, filtering out noise and ambiguities typical of raw log data. We demonstrate the usefulness of the \glspl{kg} through a final, \gls{llm}-based classification step where the objective is to map the generated \glspl{kg} to \mitreattack~\cite{stromMitreAttckDesign2018}, an established knowledge base of adversary tactics and techniques based on real-world observations. We evaluate OntoLogX by extracting structured intelligence from both a public log dataset, thereby ensuring reproducibility and comparability, as well as a new honeypot dataset that represents real-world adversarial activity. Extracted intelligence is stored in an ontology-enriched graph database, supporting semantic exploration and downstream \gls{cti} applications. We publicly release the code\footurl{https://github.com/LucaCtt/ontologx} and datasets\footurl{https://zenodo.org/records/17251494} used in this project.

In summary, we propose the following key contributions:
\begin{enumerate}
    \item A lightweight ontology and SHACL schema tailored for cybersecurity logs;
    \item OntoLogX, a retrieval-augmented, llm-based \gls{ai} agent that transforms raw log events into ontology-compliant \glspl{kg}.
    \item An ablation study of the various components of OntoLogX, showing the increased accuracy of the methodology across various \glspl{llm}.
    \item Empirical evaluation of the effectiveness of OntoLogX in extracting \gls{cti} from a real-world honeypot log dataset.
\end{enumerate}

The remainder of this paper is organized as follows. \Cref{sec:background} reviews the background concepts underlying the proposed approach, including \glspl{llm}, \gls{ai} agents, and ontologies, and discusses relevant work in \gls{cti} extraction and log analysis. \Cref{sec:methodology} presents the design of OntoLogX, detailing its ontology, \gls{rag} pipeline, validation process, and tactics prediction module. \Cref{sec:experiments} reports the experimental evaluation, covering both \gls{kg} generation and \mitreattack\ tactics prediction. \Cref{sec:conclusions} concludes the paper, summarizing key findings and outlining directions for future research.

\section{Background}\label{sec:background}
This section provides the conceptual and technical background underlying OntoLogX. We first review advances in \glspl{llm} and \gls{rag}, which form the foundation of the system's generative and grounding capabilities (\Cref{sec:llm_rag}). We then outline the notion of autonomous AI agents to situate our design within recent literature (\Cref{sec:ai_agents}). Finally, we discuss cybersecurity ontologies and validation mechanisms, which constitute the semantic layer of OntoLogX (\Cref{sec:ontologies}), before reviewing related systems in \gls{cti} (\Cref{sec:related_work}).

\subsection{Large Language Models and RAG}\label{sec:llm_rag}

\Glspl{llm} are machine learning models, usually transformer-based, that are trained on massive text corpora to carry out a variety of tasks related to natural language generation and understanding~\cite{vaswaniAttentionAllYou2023,devlinBERTPretrainingDeep2019,brownLanguageModelsAre2020}. By learning probabilistic representations of language, they can complete, summarize, translate, and interpret text across multiple domains, and have shown strong performance in zero-shot and few-shot scenarios~\cite{srivastavaImitationGameQuantifying2023}. Widely adopted examples include GPT~\cite{radfordImprovingLanguageUnderstanding2018}, Llama~\cite{touvronLlamaOpenEfficient2023}, Qwen~\cite{baiQwenTechnicalReport2023}, and Mistral~\cite{jiangMistral7B2023}.

Despite their versatility, \glspl{llm} do not inherently guarantee factual consistency, structural coherence, or domain-specific accuracy. Outputs may inherit biases from training data or lack sufficient grounding in external knowledge, which is especially problematic in specialized domains such as cybersecurity, where precise terminology and contextual interpretation are critical.

A common strategy to address these limitations is \Gls{rag}, which combines language generation with information retrieval to enhance factual grounding~\cite{lewisRetrievalaugmentedGenerationKnowledgeintensive2020}. In this paradigm, a retriever identifies documents relevant to a query from a knowledge base, and the retrieved content is provided to the model as an additional context. This improves factual grounding and domain relevance, particularly in areas where training data is insufficient or outdated~\cite{izacardLeveragingPassageRetrieval2021,guuRetrievalAugmentedLanguage2020}. In cybersecurity, \gls{rag} has the potential to enhance information extraction from \gls{cti} sources, where background knowledge is often necessary to interpret incomplete or ambiguous entries.

Another popular paradigm is the use of structured output, which consists in having \glspl{llm} output its results in a predefined schema, such as JSON, XML or through a function call~\cite{liuWeNeedStructured2024,liuAreLLMsGood2024}. Applying output constraints could not only streamline the currently repetitive process of developing, testing, and integrating \gls{llm} prompts for developers, but also enhance the user experience of \gls{llm}-powered features and applications~\cite{liuWeNeedStructured2024}. The traditional way for \glspl{llm} to output a constrained schema is through prompt engineering, but recent models and developer-oriented libraries have been introducing explicit functionalities for this purpose.

\glspl{llm} have been increasingly applied in cybersecurity scenarios, enhancing both defensive and offensive capabilities~\cite{zhangMGTEGeneralizedLongcontext2024}. For example, in \gls{cti} extraction, \glspl{llm} can automatically identify and normalize indicators of compromise, attack patterns, and \mitreattack\ tactics or techniques from unstructured reports or databases~\cite{zhangAttacKGBoostingAttack2025,mitraLocalIntelGeneratingOrganizational2025,boffaLogPrecisUnleashingLanguage2024}. For vulnerability analysis, \glspl{llm} assist in exploit explanation, vulnerability classification, and code auditing, reducing analyst workload while improving interpretability~\cite{zhouLargeLanguageModel2024,tambergHarnessingLargeLanguage2025}. They also support malware analysis and detection, where models like GPT have been used to identify malicious code, identify bugs, and obfuscated payloads~\cite{yanPromptEngineeringassistedMalware2025,tianDebugBenchEvaluatingDebugging2024}. Beyond defensive applications, \glspl{llm} have been explored for offensive purposes, such as generating phishing emails, crafting social engineering content, or simulating adversarial reasoning for red teaming~\cite{gioacchiniAutoPenBenchVulnerabilityTesting2025, barrettIdentifyingMitigatingSecurity2023,sharmaImpactBigData2023}. Although structured outputs and retrieval grounding improve factuality, they do not ensure semantic consistency of the generated output.

\subsection{AI Agents}\label{sec:ai_agents}

AI agents are autonomous software entities engineered to perform goal-directed tasks within bounded digital environments~\cite{acharyaAgenticAIAutonomous2025,sadoExplainableGoaldrivenAgents2023}. While the concept has gained renewed popularity with the rise of \glspl{llm}, definitions of autonomous agents date back decades~\cite{castelfranchiModellingSocialAction1998,franklinItAgentJust1997}.\ In general, an autonomous agent is understood as a system situated in an environment, capable of perceiving and acting on it over time in pursuit of defined objectives, and so as to effect what it senses in the future~\cite{franklinItAgentJust1997}.

\gls{ai} agents are generally characterized by three properties~\cite{sapkotaAIAgentsVs2025}: (i) \textit{autonomy}, the ability to operate with minimal human intervention; (ii) \textit{task-specificity}, a focus on narrow, well-defined goals; and (iii) \textit{reactivity and adaptability}, allowing them to respond to real-time inputs, learn from interactions, and adjust their behavior. These traits distinguish agents from deterministic automation scripts, which follow rigid workflows, as well as from stand-alone \glspl{llm}, that mainly act as reactive prompt followers.

Finally,\ \cite{sapkotaAIAgentsVs2025} reviews four use cases where AI agents are commonly used: (i) customer support automation and internal enterprise search; (ii) email filtering and prioritization; (iii) personalized content recommendation and basic data reporting; and (iv) autonomous scheduling assistants. While non-exhaustive of their capabilities, these applications highlight the potentialities of AI agents in a wide range of fields.

\subsection{Ontologies}\label{sec:ontologies}

Ontologies are formal, explicit specifications of shared conceptualizations, including classes, relationships, and constraints within a domain~\cite{staabKnowledgeProcessesOntologies2001}. They often serve as schema, guiding the construction of \glspl{kg}, and ensuring consistency and semantic clarity.

The quality and compliance of ontology-based \glspl{kg} can be assessed through \gls{shacl}~\cite{knublauchShapesConstraintLanguage2017}.\ \gls{shacl} provides a declarative framework for defining and enforcing constraints on types, property cardinalities, and relationship patterns. Such validation ensures that automatically generated knowledge remains consistent, explainable, and auditable, even when derived from noisy or incomplete data~\cite{xueKnowledgeGraphQuality2023,rabbaniSHACTORImprovingQuality2023}. This in turn enhances the robustness of downstream \gls{cti} tasks.

In cybersecurity, ontology-based frameworks are increasingly adopted to organize and standardize threat-related knowledge, enabling semantic interoperability, automated reasoning, and improved information integration across heterogeneous sources~\cite{syedUCOUnifiedCybersecurity2016}. Several well-known ontologies exist, each targeting different needs.\ \gls{uco}~\cite{syedUCOUnifiedCybersecurity2016} provides comprehensive concepts and relationships for broad cybersecurity knowledge integration.\ \gls{stix}~\cite{barnumStandardizingCyberThreat2014} defines a widely used standard for threat intelligence exchange. CRATELO~\cite{oltramariBuildingOntologyCyber2014} focuses on cyber incident and forensic data.\ \gls{misp}~\cite{wagnerMISPDesignImplementation2016} targets structured sharing of malware and threat indicators. Particularly relevant to this work is the SEPSES ontology~\cite{kieslingSEPSESKnowledgeGraph2019}, which provides a vocabulary for integrating already-parsed logs but does not support information extraction from free-text messages or interaction with language models.

Although these ontologies provide solid conceptual foundations, they were not originally conceived to guide or validate the outputs of retrieval-augmented or generative models. In the following section we examine how recent systems have attempted to bridge these gaps and position OntoLogX within this evolving landscape.

\subsection{Related Work}\label{sec:related_work}

A range of methods has been proposed for analyzing logs and extracting \gls{cti}, both using traditional text-based and \gls{llm}-based approaches.

Log parsing, defined as the process of dividing logs into static parts (static messages) and dynamic parts (variables)~\cite{maLibreLogAccurateEfficient2024}, is often used as a preliminary step. While efficient and capable of online processing, parsing alone does not impose a standardized structure, limiting subsequent reasoning. SLOGERT~\cite{ekelhartSLOGERTFrameworkAutomated2021} extended log parsing by constructing a \gls{kg} in \gls{rdf} based on a custom ontology, enabling continuous integration of parsed logs into an explorable, queryable graph that integrates multiple log sources. KRYSTAL~\cite{kurniawanKRYSTALKnowledgeGraphbased2022} built on this approach by combining declarative SPARQL queries with backward-forward chaining to detect attack patterns, outputting attack graphs aligned with \mitreattack\ tactics and techniques. However, we argue that the variability of log formats across applications or even across versions of the same system limits the effectiveness of rule-based parsing for \gls{cti} extraction.

More recent work has explored the use of language models. LogPrécis~\cite{boffaLogPrecisUnleashingLanguage2024} fine-tuned models on small sets of labeled attacks to generate attack fingerprints aligned with \mitreattack, reducing large volumes of logs into more compact and interpretable patterns. Yet, LogPrécis lacked semantic grounding and required pre-processed log sessions rather than operating directly on raw logs. CyKG-RAG~\cite{kurniawanCyKGRAGKnowledgegraphEnhanced2024} combined rule-based and \gls{llm}-based approaches to construct \glspl{kg} from cybersecurity data, integrating symbolic queries with vector similarity search for hybrid retrieval. This methodology proved effective for synthesizing responses to user queries, but it still depended on rule-based steps for \gls{kg} construction and did not perform autonomous log analysis.

\begin{table}
    \centering
    \small
    \caption{Comparison of related work for ontology-guided knowledge extraction in \gls{cti}.}
    \begin{tabular}{lccccc}
        \toprule
        \textbf{Work / Ontology}                                   & \textbf{Domain}     & \textbf{Retrieval} & \textbf{Ontology} & \textbf{Validation} & \textbf{LLM-based} \\
        \midrule
        UCO~\cite{syedUCOUnifiedCybersecurity2016}                 & Cyber ontology      & \xmark             & \cmark            & \cmark              & \xmark             \\
        STIX~\cite{barnumStandardizingCyberThreat2014}             & Threat intel        & \xmark             & \cmark            & \cmark              & \xmark             \\
        CRATELO~\cite{oltramariBuildingOntologyCyber2014}          & \gls{cti} ontology  & \xmark             & \cmark            & \cmark              & \xmark             \\
        MISP~\cite{wagnerMISPDesignImplementation2016}             & Threat sharing      & \xmark             & \cmark            & \cmark              & \xmark             \\
        SEPSES~\cite{kieslingSEPSESKnowledgeGraph2019}             & Security logs       & \xmark             & \cmark            & \cmark              & \xmark             \\
        SLOGERT~\cite{ekelhartSLOGERTFrameworkAutomated2021}       & Security logs       & \xmark             & \cmark (custom)   & \cmark (manual)     & \xmark             \\
        KRYSTAL~\cite{kurniawanKRYSTALKnowledgeGraphbased2022}     & \gls{cti} reports   & \xmark             & \cmark            & \xmark              & \xmark             \\

        CyKG-RAG~\cite{kurniawanCyKGRAGKnowledgegraphEnhanced2024} & \gls{cti} documents & \cmark             & \xmark            & \xmark              & \cmark             \\
        \midrule
        OntoLogX (this work)                                       & System logs         & \cmark             & \cmark            & \cmark (SHACL)      & \cmark             \\
        \bottomrule
    \end{tabular}\label{tab:related-work}
\end{table}

Table~\ref{tab:related-work} summarizes representative systems and ontologies in \gls{cti}.\ Prior works addressed individual aspects such as ontology alignment or retrieval grounding, OntoLogX instead integrates retrieval, ontology guidance, and validation, within a single \gls{llm}-based framework.

\section{OntoLogX Framework Design}\label{sec:methodology}
This section details the architecture and operational workflow of OntoLogX.
We describe how the system integrates retrieval, generation, and validation modules under an ontology-guided framework to transform raw log data into structured, semantically consistent knowledge graphs.
Each subsection outlines the key components their interactions, and the design of the underlying ontology used to structure the knowledge extracted from logs.

\subsection{Overview}\label{sec:overview}

OntoLogX is an \gls{ai} agent for online log analysis, designed to process events incrementally and sequentially, one by one, in a setting that reflects realistic cybersecurity use cases requiring near real-time event analysis. Each log event is analyzed with the support of an \gls{llm}, which produces an ontology-grounded \gls{kg} representation. Optional context information (e.g., device, process, operating system, honeypot version) can be provided to enrich the generation process, even if unstructured.

The proposed agent does not aim to reconstruct complete attack narratives at the event level; instead, each KG represents a minimal, self-contained representation of an event grounded in the log ontology. Even when a log line captures only a fragment of activity, it typically contains entities (e.g., source, application, user, network address, timestamp) and relationships that can be represented explicitly. In realistic deployment scenarios, logs arrive sequentially and must be processed without assuming access to future events. Producing one KG per event enables immediate storage, retrieval, and reuse in the RAG component, and avoids the need for buffering or session reconstruction during generation. Higher-level reasoning over multiple events can be handled in a later stage of the pipeline.

\begin{figure}
    \centering
    \includegraphics[width=0.8\linewidth]{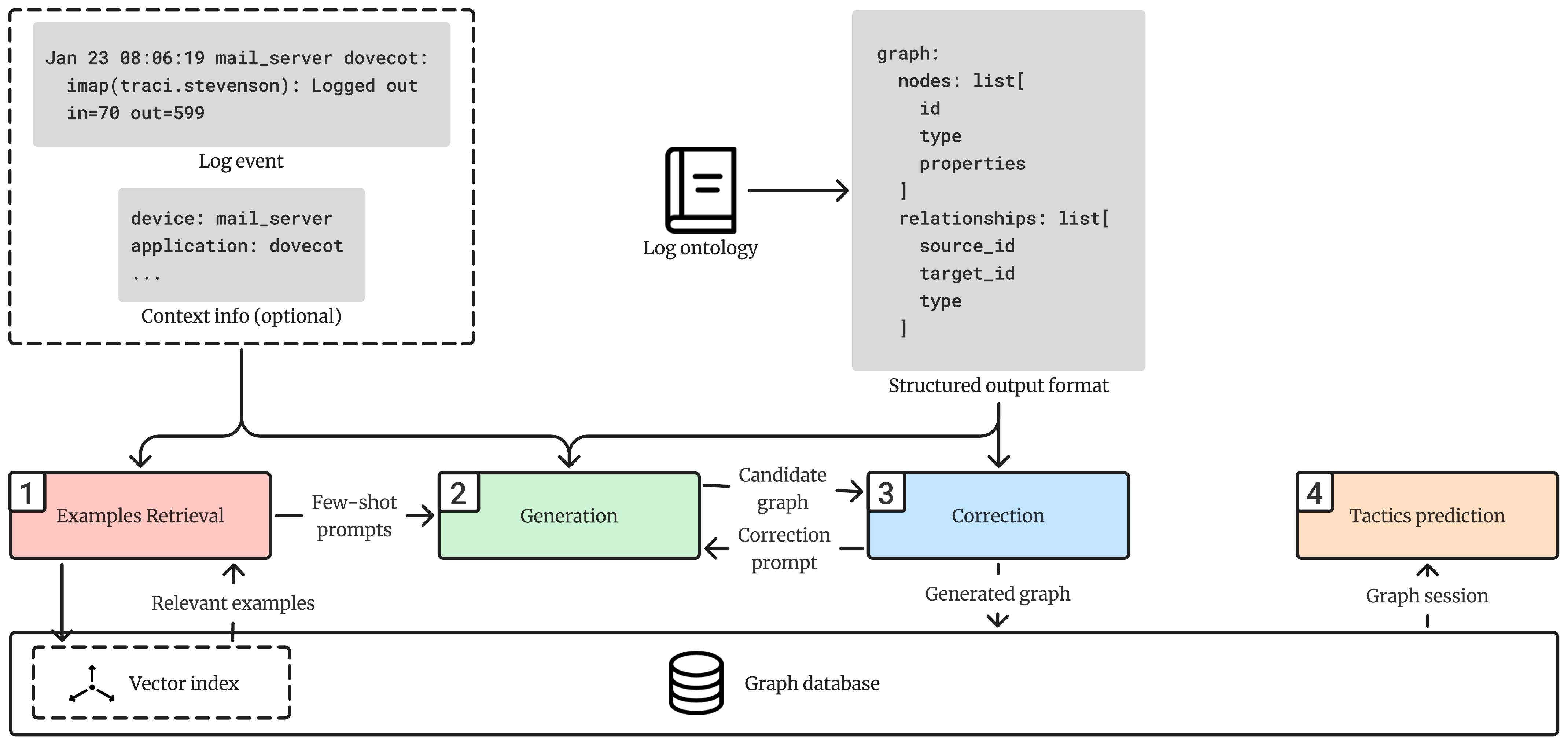}
    \caption{Methodology for generating a log event \gls{kg}, starting from the raw log event and optional context information.}\label{fig:methodology}
\end{figure}

\Cref{fig:methodology} illustrates the overall workflow. When a log event arrives, the system first retrieves semantically related log event \glspl{kg} from the graph database. These serve as few-shot examples to help the \gls{llm} adapt its output to the ontology and to previously seen patterns. The \gls{llm} then generates a candidate \gls{kg} by combining the new log event, the optional context, and the domain ontology. The candidate is validated against ontology constraints: if the output is malformed or non-compliant, the model is prompted again within the same interaction to apply targeted corrections. This iterative refinement continues until a valid representation is obtained. Once validated, the \gls{kg} is stored independently in the graph database, ensuring that it can be retrieved for future processing without requiring immediate integration with other graphs. Finally, the generated \glspl{kg} are grouped depending on the log session they originate from, and each session is used to predict associated \mitreattack\ tactics labels through an \gls{llm} call. It is worth noting that \glspl{kg} are stored independently from each other, as the semantic connection of different \glspl{kg} is not the focus of this work.

A key element enabling this process is the underlying log ontology.
The following subsection describes the structure and rationale of this ontology, detailing how it balances expressiveness and simplicity to support automated extraction from raw logs.

\subsection{Ontology Design for Structured Log Representation}\label{sec:log-ontology}

The \glspl{kg} generated by OntoLogX are grounded in a custom log ontology that formalizes information extracted from raw events. Besides providing structure, the ontology guides the \gls{llm} during generation by indicating which elements to identify in each log. This is particularly important because log entries, whether structured or unstructured, often contain significant information but may encode it inconsistently or without explicit separation. %\Cref{lst:log-examples} reports examples for both types of logs, which capture information such as timestamps, user identities, network addresses, and certificate details.

We argue that existing cybersecurity ontologies are not well-suited for \gls{llm}-based log processing. Minimal models, such as the one in SLOGERT's, capture too few concepts to be useful for \gls{cti} analysis. In contrast, large frameworks like \gls{uco} are overly complex for automated generation: their size increases the likelihood of errors, and they assume pre-parsed or structured metadata, which is rarely available in raw logs. To address these issues, OntoLogX employs a novel, lightweight, yet expressive ontology tailored to log characteristics. Using a predefined ontology also ensures formalization and consistency, while the methodology itself remains flexible: thanks to the \gls{llm}-driven pipeline, the ontology can be swapped or extended as needed by the user.

\begin{figure}
    \centering
    \includegraphics[width=0.8\linewidth]{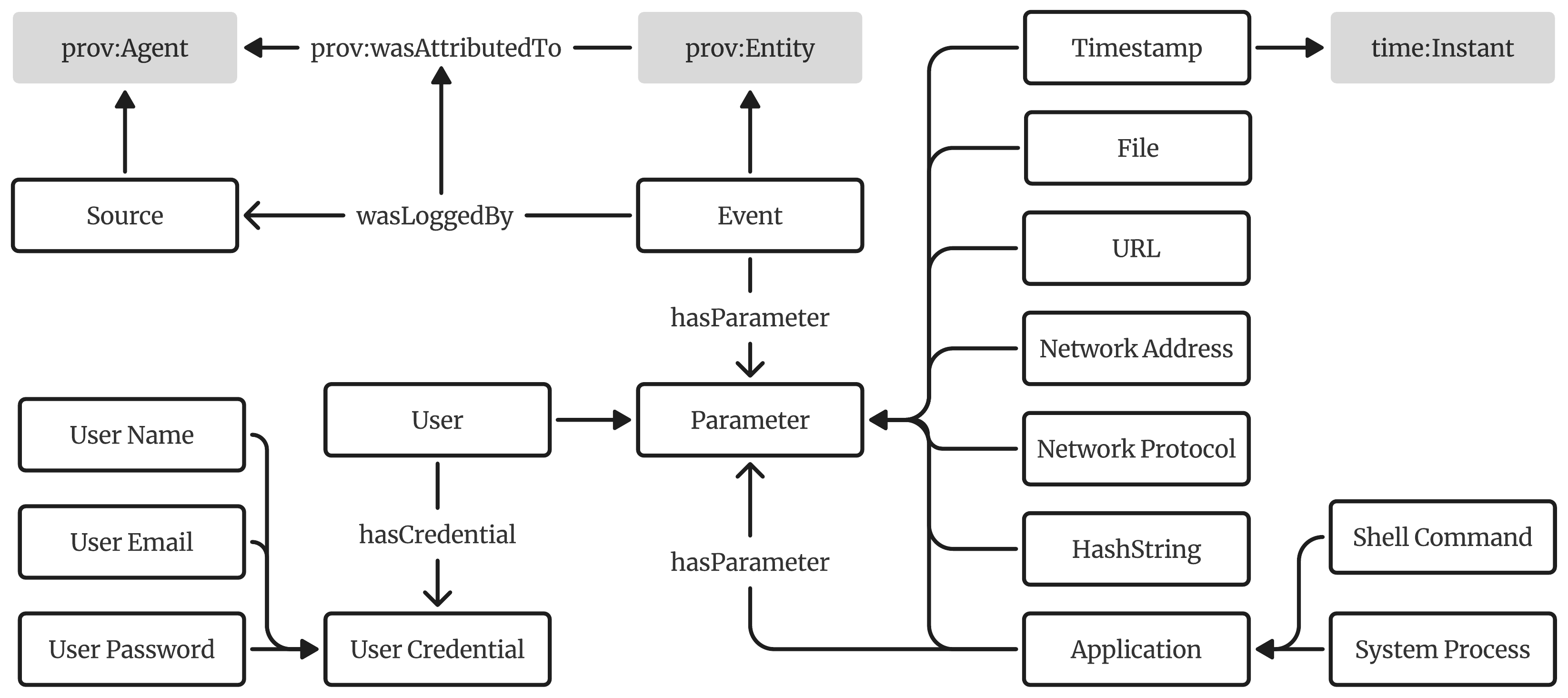}
    \caption{Classes and object properties of the OntoLogX ontology. Data properties are omitted for conciseness. Full arrows indicate either \texttt{rdfs:subClassOf} or \texttt{rdf:subPropertyOf} object properties. Gray boxes highlight classes from external ontologies.}\label{fig:ontology}
\end{figure}

The schema, shown in \Cref{fig:ontology}, is designed to capture the most common concepts in cybersecurity logs without being rigid or monolithic. At its core is the \texttt{Event} class, which represents a single log entry. Each event is linked to a \texttt{Source}, describing the device or application that produced the log. These two classes are mapped to the \texttt{Entity} and \texttt{Agent} classes in the \texttt{prov-o} ontology, aligning them with provenance standards. The information contained within the logs themselves is represented as subclasses of \texttt{Parameter}, including a dedicated \texttt{TimeStamp} parameter aligned with the W3C \texttt{time} ontology. More complex structures are also supported: the \texttt{Application} parameter can reference other parameters, enabling the modeling of application call arguments or chains of calls. \texttt{UserCredential} models various credentials that a \texttt{User} can have, with specialized classes available for username, email, and password (for more details, see Appendix~\ref{app:ontology}).

To ensure quality and compliance, we developed a companion \gls{shacl} specification. These constraints enforce schema validity by checking property cardinalities, type consistency, and the presence of required fields. This validation step is especially valuable in an \gls{llm}-based pipeline, where outputs may otherwise be incomplete or inconsistent. Together, the ontology and its constraints guarantee that generated \glspl{kg} remain semantically coherent, queryable, and interoperable with broader \gls{cti} frameworks.

While the ontology provides the structural and semantic blueprint for knowledge representation, OntoLogX further enhances generation quality through contextual grounding.
Before producing a new graph, the system retrieves semantically similar examples that illustrate how comparable events were previously represented.
The next subsection outlines this hybrid retrieval mechanism.

\subsection{Examples Retrieval}\label{sec:examples-retrieval}

OntoLogX incorporates an example retrieval step, which guides the \gls{llm} in constructing \glspl{kg} and enables reuse of knowledge from related logs. The objective is to identify semantically and textually similar log entries that can serve as few-shot prompts, thereby improving both the structure and consistency of the generated graphs. Examples are drawn from a dedicated store indexing previously generated \glspl{kg} as well as manually annotated instances aligned with the log ontology.

\begin{figure}
    \centering
    \includegraphics[width=0.8\linewidth]{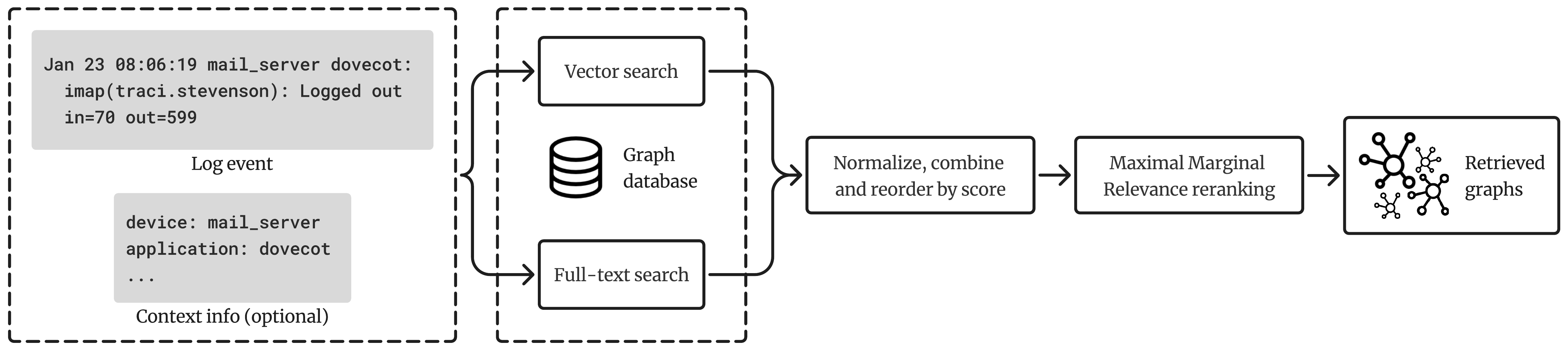}
    \caption{Hybrid retrieval process.}\label{fig:retrieval}
\end{figure}

As illustrated in \Cref{fig:retrieval}, retrieval is performed through a hybrid strategy combining vector and full-text search. The input log and its context are queried against both indices, allowing for semantic similarity matching and precise word-based lookups. The vector index stores embeddings of the raw log event and context information used to generate the graph. The full-text index, instead, contains the individual words of the log event and its context. Using only one of these approaches would be insufficient: full-text search alone misses semantic nuances that help the model capture hidden relationships, while vector search alone risks overlooking near-identical matches, which are often the most useful as generation examples. The results from each search are divided by the maximum score for that search type, e.g, vector search scores are divided by the maximum vector search score. Then the results from both searches are combined and sorted in decreasing order by their normalized score.

OntoLogX finally employs \gls{mmr}~\cite{carbonellUseMMRDiversitybased1998}, a re-ranking strategy that balances \textit{relevance} with \textit{diversity}. Rather than returning only the top-\(k\) most similar items, \gls{mmr} penalizes redundancy by favoring candidates that are both close to the query and dissimilar from each other. Given a query \(q\), candidate set \(D\), and already selected items \(S \subset D\), the next example \(d^*\) is chosen by maximizing:
\[
    \text{MMR}(d) = \lambda \cdot \text{Sim}(d, q) - (1 - \lambda) \cdot \max_{s \in S} \text{Sim}(d, s)
\]
where \(\text{Sim}(d, q)\) measures the similarity between document \(d\) and query \(q\), \(\text{Sim}(d, s)\) measures the similarity between \(d\) and already selected document \(s\), and \(\lambda \in [0,1]\) controls the trade-off between relevance and diversity. Using \gls{mmr} helps ensure that the retrieved examples cover a broader range of patterns and structures, rather than being clustered around a single interpretation of the log event.

The retrieved examples and the input log are then combined to form the prompt for the large language model.
This step anchors the model to both the ontology structure and relevant prior instances, ensuring coherence and consistency in the generated knowledge graphs.
The following subsection details this ontology-guided generation process.

\subsection{Ontology-Guided KG Generation with LLMs}\label{sec:generation}

\begin{figure}
    \centering
    \includegraphics[width=0.8\linewidth]{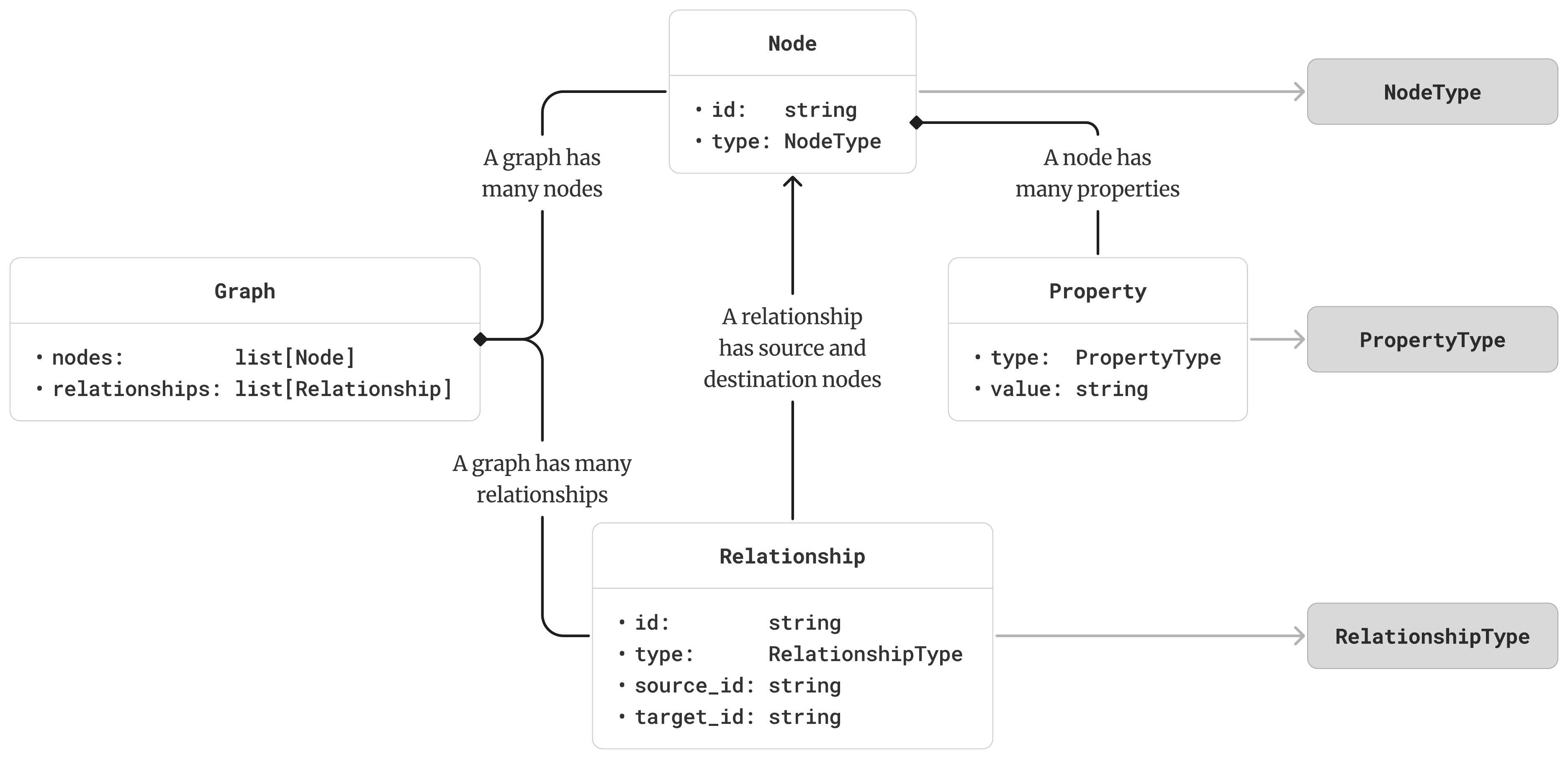}
    \caption{Format of structured output. \texttt{NodeType}, \texttt{PropertyType}, and \texttt{RelationshipType} respectively represent the valid classes, data properties, and object properties defined in the ontology.}\label{fig:structured-output}
\end{figure}

\begin{listing*}
    \begin{minted}
[
frame=lines,
framesep=2mm,
baselinestretch=1,
fontsize=\scriptsize,
breaklines=true,
breaksymbolleft={},
breaksymbolright={}
]
{turtle}
@prefix : <https://cyberseclab.unibs.it/olx/examples/#>.
@prefix olx: <https://cyberseclab.unibs.it/olx/dict#>.
@prefix rdf: <http://www.w3.org/1999/02/22-rdf-syntax-ns#>.
@prefix time: <http://www.w3.org/2006/time#>.
@prefix xsd: <http://www.w3.org/2001/XMLSchema#>.

:timestamp-3 rdf:type olx:TimeStamp;
  time:inXSDDateTimeStamp "2022-01-20T13:47:21Z"^^xsd:dateTimeStamp;

:system-process-3 rdf:type olx:SystemProcess;
  olx:applicationName "dnsmasq";
  olx:systemProcessPID 3326.

:network-address-3 rdf:type olx:NetworkAddress;
  olx:networkAddressHostname "d1zkz3k4cclnv6.cloudfront.net";
  olx:networkAddressIPV4 "192.168.231.180".

:source-3 rdf:type olx:Source;
  olx:sourceName "dnsmasq";
  olx:sourceDevice "inet-dns".

:event-3 rdf:type olx:Event;
  olx:hasParameter :timestamp-3, :system-process-3, :network-address-3;
  olx:wasLoggedBy :source-3.
\end{minted}
    \caption{Example of a \gls{kg} produced by OntoLogX for the log event \texttt{Jan 20 13:47:21 dnsmasq[3326]: query[AAAA] d1zkz3k4cclnv6.cloudfront.net from 192.168.231.180} from the \texttt{dnsmasq} application running on the \texttt{inet-dns} device.}\label{lst:kg-example}
\end{listing*}

In the generation step, OntoLogX integrates a \gls{llm} to produce a \gls{kg} from each log event and its associated context. The model is also guided by the relevant \glspl{kg} retrieved in the previous step, which serve as few-shot examples. This combined input both anchors the model to the desired output structure and provides historical context from semantically and textually related logs, improving the coherence and utility of the generated graphs. The prompt used to instruct the \gls{llm} (see \Cref{lst:main-prompt} in the Appendix) is designed to be model-agnostic, maximizing compatibility with a broad range of language models. It comprises: (i) a clear role and task definition, (ii) detailed instructions grounded in the OntoLogX ontology, and (iii) a set of constraints and clarifications based on common failure patterns observed in early experiments, such as malformed URIs, incorrect predicate usage, and mismatched types. The expected ontology format is enforced through the use of the structured output schema shown in \Cref{fig:structured-output}. This format defines the required fields and their expected types, effectively serving as a strong constraint during generation and simplifying validation. \ \Cref{lst:kg-example} provides an example of a generated KG in turtle format, converted from the structured output.

A key advantage of using a \gls{llm} is its ability to generate high-quality \glspl{kg} without requiring domain-specific supervised training. By leveraging knowledge gained during pretraining, the model can infer implicit information, disambiguate vague or underspecified log entries, and normalize inconsistent terminology. We argue that these capabilities are particularly useful for cybersecurity logs, where entities and activities are often expressed in non-standard, abbreviated, or noisy forms. The generalization afforded by pretrained \glspl{llm} enables the system to handle heterogeneous sources and adapt to evolving logging formats.

Although ontology guidance significantly constrains the generation process, LLM outputs can still contain structural or semantic inconsistencies.
To guarantee compliance with the defined schema and maintain the integrity of the knowledge base, OntoLogX performs an explicit validation and correction phase. This stage verifies and refines the generated graphs through an iterative SHACL-based feedback loop, as described next.

\subsection{Iterative Validation and Correction with SHACL Rules}\label{sec:correction}

A drawback in using \glspl{llm} for ontology-grounded \gls{kg} generation is the risk of producing outputs that are incomplete, malformed, or semantically inconsistent with the intended schema. To mitigate this, OntoLogX introduces a dedicated \textit{correction} phase, where generated graphs are automatically validated and, if necessary, revised through iterative feedback with the model.

The correction pipeline proceeds in three stages. First, the syntactic validity of the output is checked to ensure that nodes and relationships are properly defined and the graph conforms to the required structured format. Second, ontology compliance is verified by enforcing the constraints specified in the SHACL rules. This step covers the correct use of classes and properties, consistent data typing, and the satisfaction of required schema rules. Third, semantic validation is performed to detect higher-level inconsistencies, such as: (i) the absence of an \texttt{Event} node, (ii) the presence of multiple \texttt{Event} nodes, (iii) relationships pointing to undefined entities, or (iv) duplicate node definitions.

When violations are identified, OntoLogX constructs a targeted correction prompt that highlights the errors and requests specific revisions from the \gls{llm}. This feedback loop can iterate across multiple rounds, progressively refining the output until a fully valid and ontology-compliant \gls{kg} is obtained. If a valid graph cannot be produced within the allowed attempts, the system defaults to outputting an empty graph, which negatively impacts evaluation but ensures that invalid results do not contaminate the knowledge base.

Once a corrected \gls{kg} passes all validation stages, it is persisted in the graph database along with the originating log event and any contextual metadata, enabling future retrieval and traceability should further analysis be required.

\subsection{Downstream Application: Tactics Prediction}\label{sec:tactics_prediction}

To illustrate this capability and to evaluate the practical benefits of OntoLogX, we implement a final classification stage that receives \gls{kg} to determine \mitreattack\ tactics associated with logs of events generated during attacks. This step provides higher-level semantic insights inferred across multiple, related events, and also helps in contextualizing log activity within a widely adopted threat intelligence framework.

The prediction process begins once individual log \glspl{kg} have been generated and validated. These graphs are first grouped into sessions, either by capturing sets of events that are temporally related or by profiting from common properties of the graphs (such as common applications). An \gls{llm} is then prompted to analyze the aggregated \glspl{kg} and assign one or more \mitreattack\ tactics that best describe the observed behavior. The prompt used for this task is reported in \Cref{lst:tactics-ontologx-prompt} in the Appendix.

Using \glspl{kg} rather than raw logs offers several advantages for this task. First, \glspl{kg} provide a normalized, ontology-compliant representation that abstracts away the heterogeneity of log formats, allowing the model to focus on entities, relationships, and temporal structure rather than noisy or inconsistent syntax. Second, the explicit structuring of events into semantic components facilitates reasoning about higher-level attack stages, making it easier to identify the behaviors of attackers. In contrast, relying directly on raw logs would require the model to both parse and interpret the event simultaneously, increasing the likelihood of errors and reducing generalizability across log sources.

\section{Experiments}\label{sec:experiments}
We organized the experimental evaluation of OntoLogX into two parts. First, we evaluated the quality of \gls{kg} generation by conducting an ablation study to assess the contribution of individual components of the pipeline (retrieval, correction, and structured output), and comparing the performance of different language models. Second, we examined the effectiveness of OntoLogX when used on the downstream tactics prediction problem, leveraging a cybersecurity-oriented \gls{llm} to map log sessions to \mitreattack\ tactics.\ We conducted the evaluation using both honeypot and a simulated dataset because they serve complementary evaluation goals. Honeypot data allows us to assess performance on real-world attacker behavior under high adversarial activity, while benchmark synthetic data ensures reproducibility, robustness and applicability to logs presenting varying formats and which may not be attack-centric.

The implementation of OntoLogX was designed to be general and model-agnostic, ensuring that the methodology does not depend on a specific \gls{llm}. For \gls{kg} generation, structured output was enforced through function-calling interfaces, which constrain the \gls{llm} to produce results in a predefined format. The correction phase was limited to a maximum of three correction steps per log event; if no valid ontology-compliant graph was produced within this limit, the output was considered an empty graph.\ This value was selected as an empirically effective trade-off between output quality and computational efficiency. In preliminary experiments, most structural and ontological violations were resolved within the first one or two iterations, while a third iteration occasionally corrected residual issues such as missing entities or incorrect property usage. Using fewer iterations resulted in avoidable SHACL violations or incomplete graphs. All validated \glspl{kg} were stored in a Neo4j\footurl{https://neo4j.com/} graph database, extended with a vector index to support semantic and full-text retrieval over node property embeddings. For embedding generation, we adopted the \texttt{gte-multilingual-base} model~\cite{zhangMGTEGeneralizedLongcontext2024}. To facilitate reproducibility and enable the structured sharing of experimental results, the database was further organized according to the MLSchema ontology~\cite{publioMLschemaExposingSemantics2018}.

\subsection{Knowledge Graph Construction}

The first experiment evaluated the effectiveness of OntoLogX in constructing ontology-compliant \glspl{kg} from raw log events. Six configurations were considered: (i) a baseline without retrieval, structured output, or correction; (ii) retrieval only; (iii) structured output only; (iv) structured output with corrections; (v) the full OntoLogX pipeline; (vi) the full pipeline with fully populated graph database and vector index for retrieval. This comparison isolates the incremental benefits of \gls{rag}, output structuring, and iterative correction mechanisms.

\subsubsection{Models}

We evaluated OntoLogX using a set of eight \glspl{llm}, selected to span different architectures, parameter scales, and licenses. The models considered along with their number of parameters are: \llamabig, \llamasmall, \claudesonnet, \claudehaiku, \mistrallarge, \gptossbig, \gptosssmall, and \qwencoder. With the exception of the Claude family, all models are distributed as open weights, making them deployable outside of proprietary cloud environments.\ \qwencoder\ is code-specialized \glspl{llm}, trained primarily on source code and related artifacts. Such models are designed to excel at generating syntactically precise and semantically consistent outputs, making them particularly well-suited for tasks involving structured representations.\ \gptossbig\ and \gptosssmall, instead, belong to a family of reasoning-oriented models that emphasize logical inference and multi-step problem solving.

All models were accessed through AWS Bedrock,\footurl{https://aws.amazon.com/bedrock/} with the exception of \qwencoder, which was executed via vLLM~\cite{kwonEfficientMemoryManagement2023} on a AWS EC2 instance equipped with four NVIDIA L4 GPUs. A temperature of 0.7 was applied to all runs and models.\ This value follows the recommended configuration for Qwen Coder and is applied uniformly to all models to avoid introducing model-specific tuning effects. To mitigate stochastic variability, we repeated each experiment ten times and considered the mean values.

\subsubsection{Dataset}

The dataset used in this evaluation step consisted of log events sampled from the AIT log dataset~\cite{landauerAITLogData2022}. A total of 70 log entries were selected to ensure both syntactic and semantic diversity. To promote heterogeneity, the first 100 events were extracted from each file in the \textit{RussellMitchell} testbed. From this pool, 70 entries were chosen using an embedding-based dissimilarity criterion. Specifically, embeddings were computed using the \texttt{nomic-embed-text-v1.5} model, and cosine distances were calculated with respect to previously selected entries. Candidates with a minimum distance below 0.7 \--- indicating excessive similarity \--- were discarded in favor of more diverse samples. Each selected log event was manually annotated with a gold-standard \gls{kg}, ensuring a reliable basis for evaluation. Finally, the dataset was randomly partitioned into three subsets: 10 examples reserved for few-shot prompting, 10 for validation during prompt refinement, and 50 for testing.

Unlike honeypot data, the AIT dataset includes logs from heterogeneous applications, devices, and formats, many of which are not attack-centric. This allows us to evaluate whether OntoLogX generalizes across diverse log structures and semantics, rather than overfitting to a single attack-focused source.

\subsubsection{Metrics}

To assess the quality of \glspl{kg} generated by OntoLogX, we combined ontological validation with semantic evaluation using \gls{llm}-based scoring:
\begin{itemize}
    \item \textit{Construction Success Ratio}: proportion of log inputs for which a non-empty \gls{kg} was successfully constructed. This metric captures the overall robustness of the pipeline.
    \item \textit{SHACL Violation Ratio}: proportion of SHACL constraints violated across generated graphs. Lower values indicate stronger adherence to the ontology’s formal rules.
    \item \textit{Precision}: fraction of generated triples (defined as \texttt{(subject, predicate, object)}, e.g., \texttt{(:event-1, rdf:type, olx:Event)}) that are correct, i.e., present in the ground-truth \gls{kg}, over the total number of generated triples. High precision reflects accurate extractions with few spurious facts.
    \item \textit{Recall}: fraction of ground-truth triples that were successfully generated, i.e., matched in the output. High recall reflects comprehensive coverage of the relevant information in the logs.
    \item \textit{F1 Score}: harmonic mean between precision and recall.
    \item \textit{Entity Linking Accuracy}: percentage of correctly generated entities, defined as class instances along with their associated properties.
    \item \textit{Relationship Linking Accuracy}: percentage of correctly generated relationships between entities that are themselves correct.
    \item \textit{G-Eval Score}: a \gls{llm}-as-a-judge framework~\cite{liuGevalNLGEvaluation2023} that employs chain-of-thought reasoning and a form-filling paradigm to evaluate natural language generation outputs. In our setting, it uses the \texttt{Llama~3.3} model to assess the semantic fidelity of generated graphs. The evaluator is prompted (see \Cref{lst:geval-prompt} in the Appendix) to produce natural language summaries of both the raw log and the \gls{kg}, and then to score their semantic overlap on a scale from 0 to 1. Additional information in the \gls{kg} reduces the score only if deemed irrelevant. Due to the inherent noisiness of logs, a higher score is not necessarily an indicator of quality: we expect that the ideal score is between 0.7--0.8, which indicates high information retention without the noise.
\end{itemize}

\subsubsection{Results}

The results of the \gls{kg} generation experiments are summarized in \Cref{fig:scores} and reported in detail in \Cref{tab:ait-results} (in the Appendix). Overall, OntoLogX proves effective in producing ontology-compliant \glspl{kg}, though the impact of individual components varies significantly across models and configurations.

\begin{figure*}
    \begin{subfigure}{.5\textwidth}
        \centering
        \caption{Construction success ratio, higher is better.}%
        \includegraphics[width=.95\linewidth]{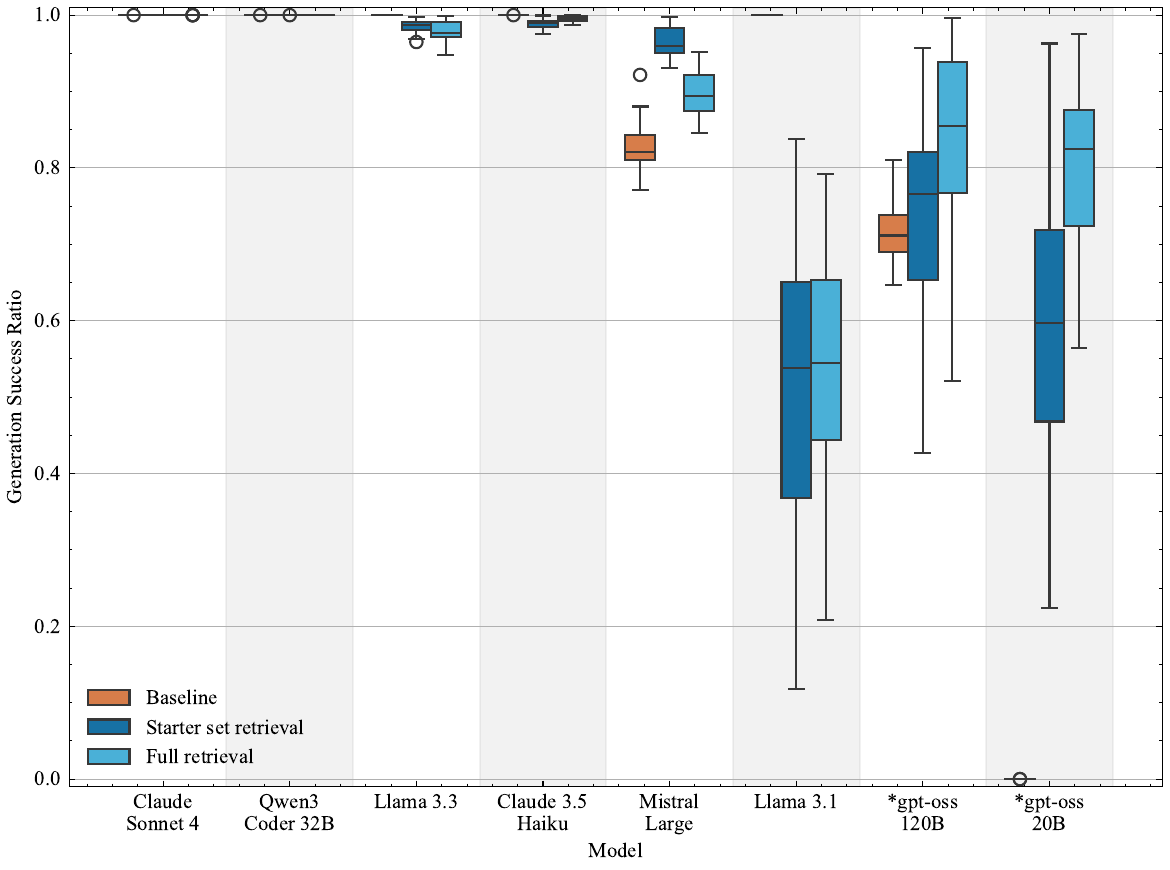}\label{fig:generation-success-ratio}
    \end{subfigure}%
    \begin{subfigure}{.5\textwidth}
        \centering
        \caption{SHACL violation ratio, lower is better.}%
        \includegraphics[width=.942\linewidth]{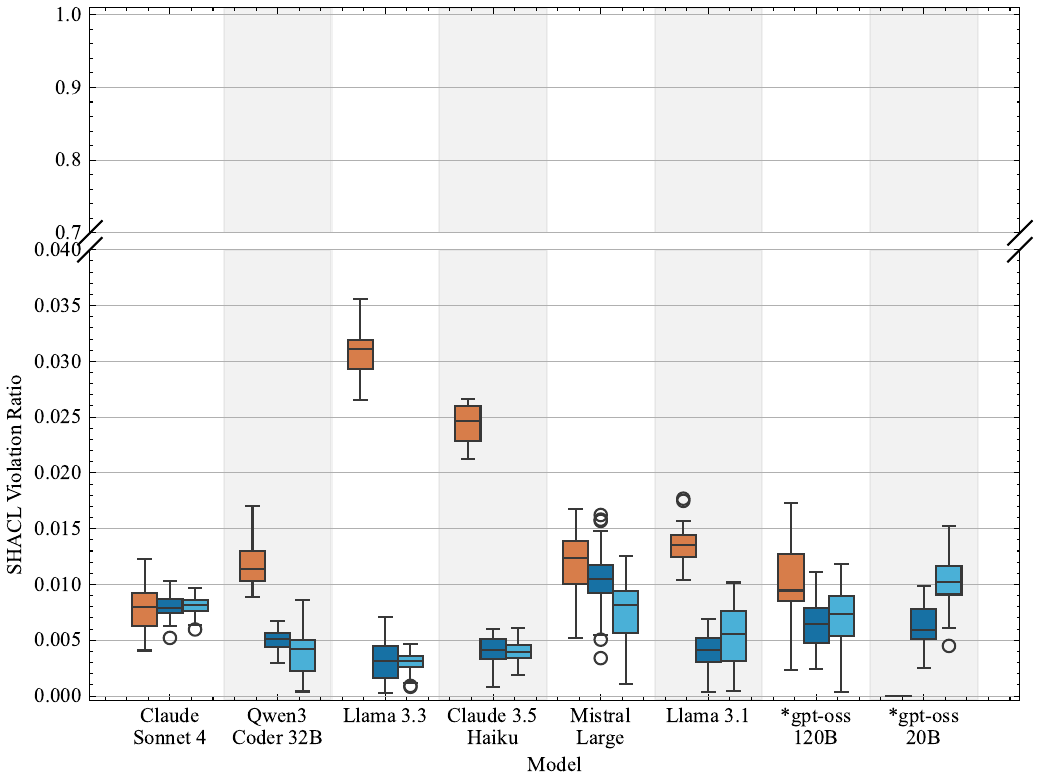}\label{fig:shacl-violation-ratio}
    \end{subfigure}
    \begin{subfigure}{.5\textwidth}
        \centering
        \caption{F1 score, higher is better.}
        \includegraphics[width=.95\linewidth]{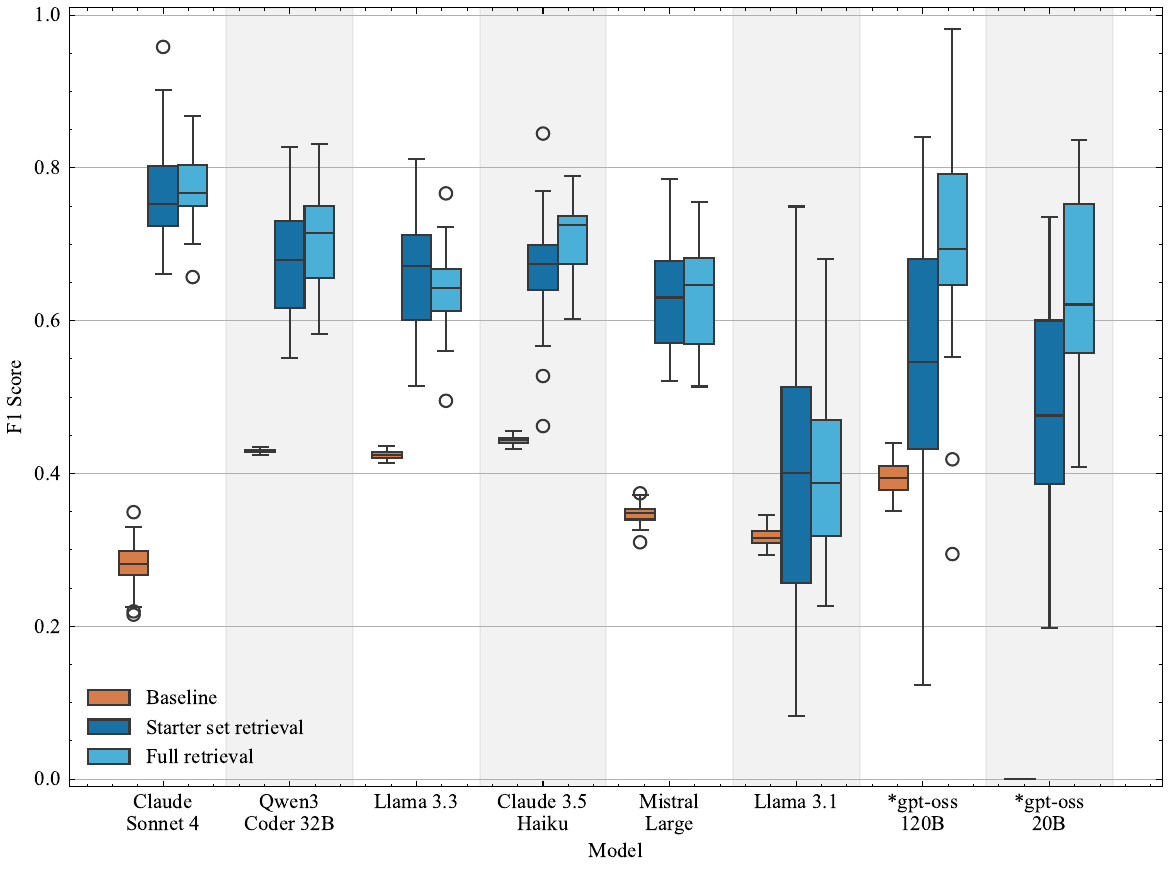}\label{fig:f1-score}
    \end{subfigure}%
    \begin{subfigure}{.5\textwidth}
        \centering
        \caption{Entity linking accuracy, higher is better.}
        \includegraphics[width=.95\linewidth]{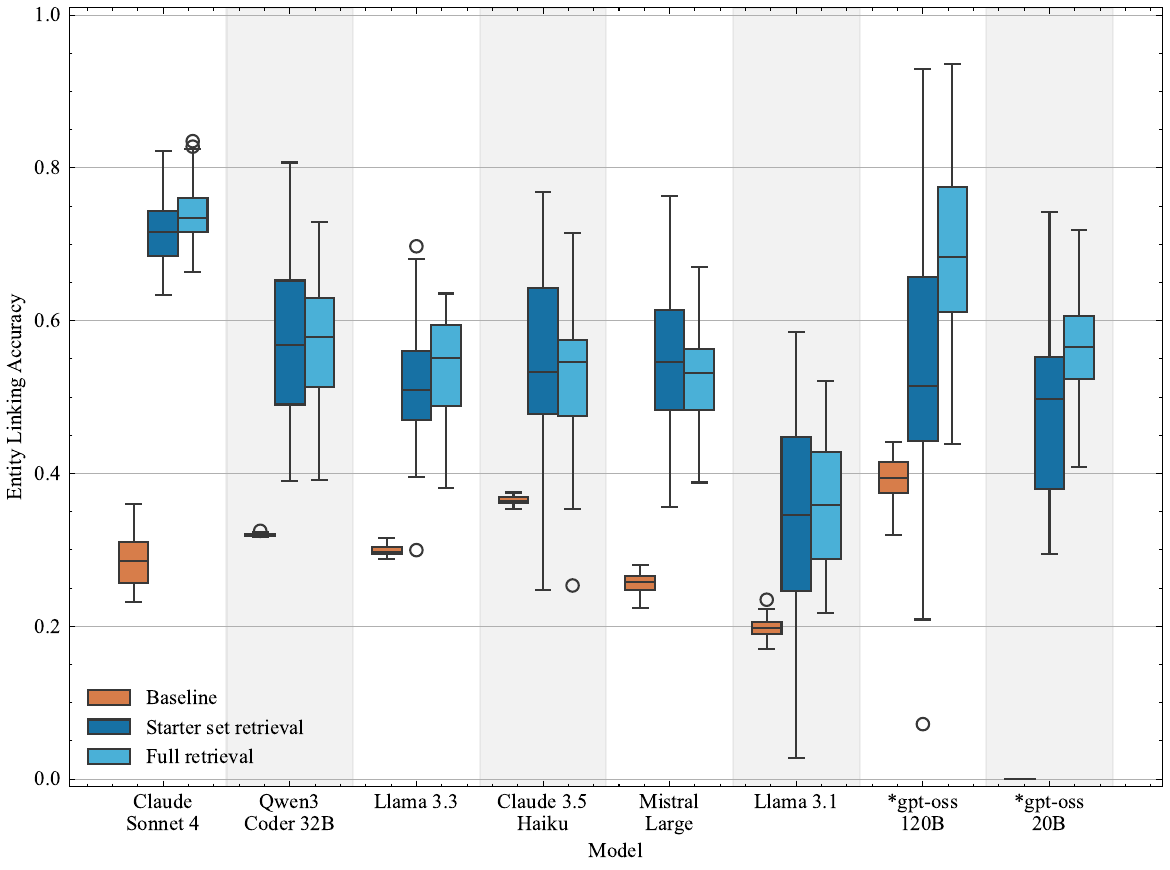}\label{fig:entity-linking-accuracy}
    \end{subfigure}
    \begin{subfigure}{.5\textwidth}
        \centering
        \caption{Relationship linking accuracy, higher is better.}%
        \includegraphics[width=.95\linewidth]{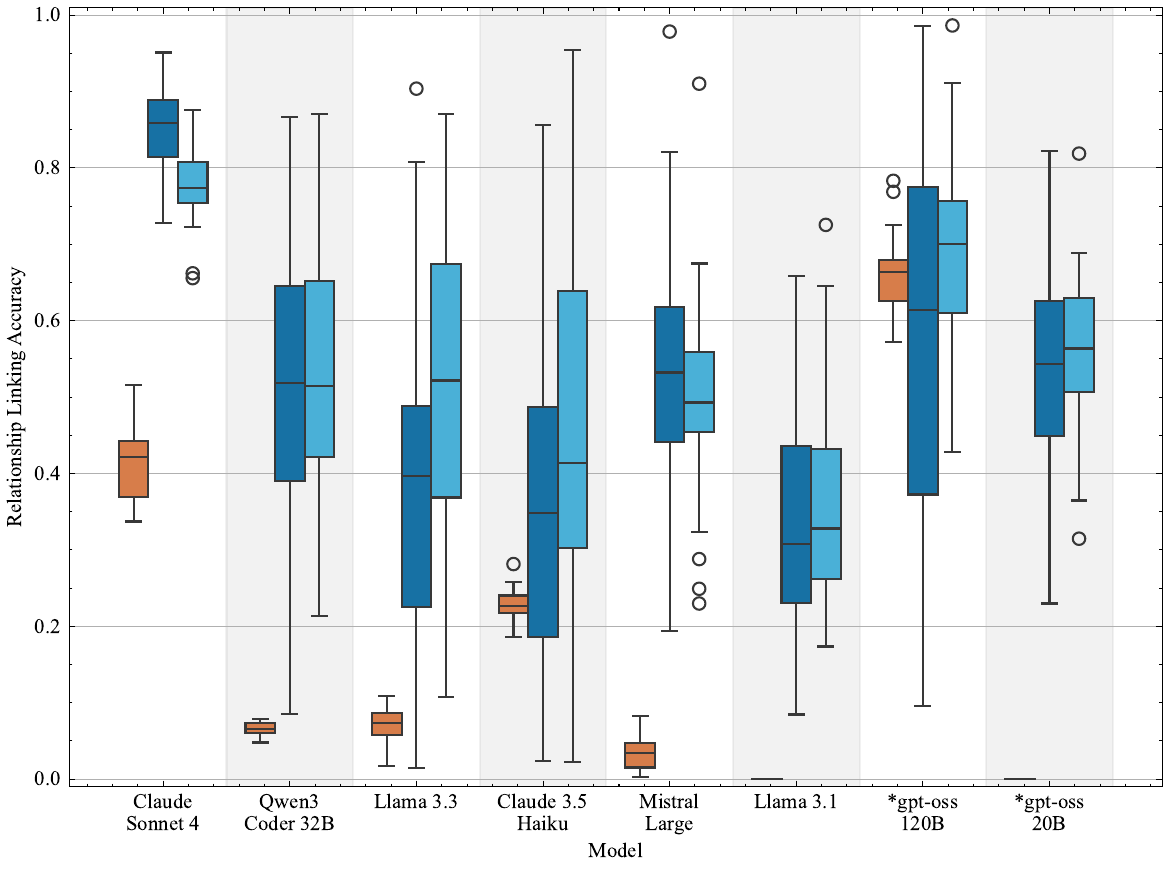}\label{fig:relationship-linking-accuracy}
    \end{subfigure}%
    \begin{subfigure}{.5\textwidth}
        \centering
        \caption{G-Eval Score.}%
        \includegraphics[width=.95\linewidth]{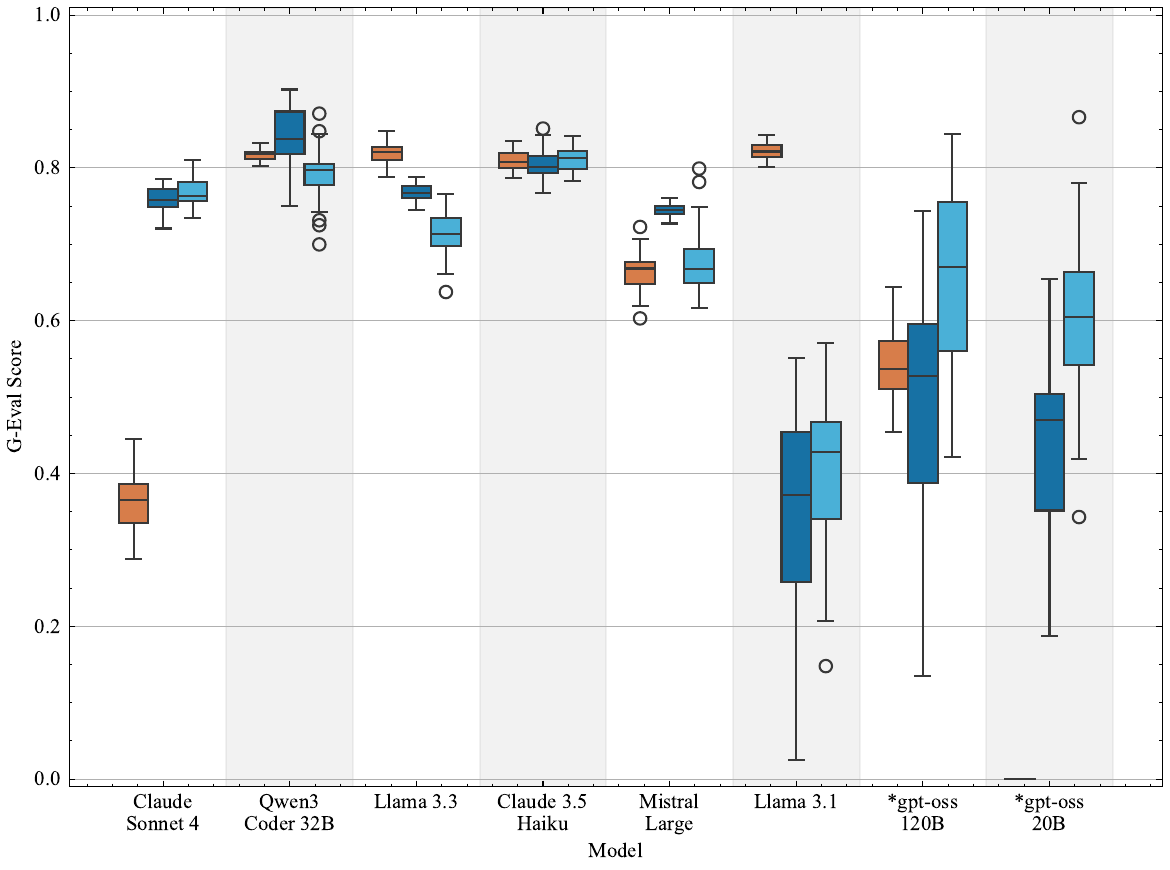}\label{fig:geval-score}
    \end{subfigure}
    \caption{Model and approach comparison. Reasoning models are highlighted with an asterisk before their name.}\label{fig:scores}
\end{figure*}

\paragraph{Construction Success and SHACL Violations.}
\claudesonnet, \qwencoder, \llamabig, and \claudehaiku\ achieve high success scores in constructing \glspl{kg}, across all configurations.\ \llamasmall\ fails often by not calling the output tool at all, instead generating free-form text or reasoning chains.\ \mistrallarge\ also struggles, by producing syntactically invalid KGs. Reasoning models in particular are discussed in detail below.

\paragraph{Reasoning Models.}
The \gptossbig\ and \gptosssmall\ reasoning models underperform across all metrics and configurations, because of their high failure rate with the structured output tool. Closer inspection reveals that these models actually do often produce valid graphs, however they call the output tool incorrectly. In particular, we observed that the invalid outputs fall into three categories:(i) partial (though syntactically correct) graph that is used within the reasoning process, rather than the final output graph it produced; (ii) malformed function calls, such as incomplete JSON objects, or unstructured text surrounding the actual arguments; (iii) outputs that completely omit function calls, instead generating free-form text or reasoning chains. These issues persist even in the presence of retrieval and correction mechanisms, leading to the conclusion that these models are not adequate for the proposed methodology, and may require alternative strategies for structured output enforcement.

\gls{shacl} violation ratios are consistently low across all setups and configurations (\Cref{fig:shacl-violation-ratio}), which suggests that prompt engineering alone can effectively guide LLMs toward generating ontologically valid structures. Nonetheless, the addition of structured output and correction mechanisms further reduces violations.

\paragraph{Execution Time.}
Direct comparison of the execution times across different models is not possible, due to differences in the underlying backends. However, within the same model, the baseline configuration consistently proves to be the fastest, reflecting its shorter prompts and thus reduced context length. The addition of retrieval, structured output, and correction mechanisms increases execution time, with each component contributing to longer processing durations.

\paragraph{Balance between G-Eval and F1 Scores.}
\begin{figure}
    \centering
    \includegraphics[width=0.5\linewidth]{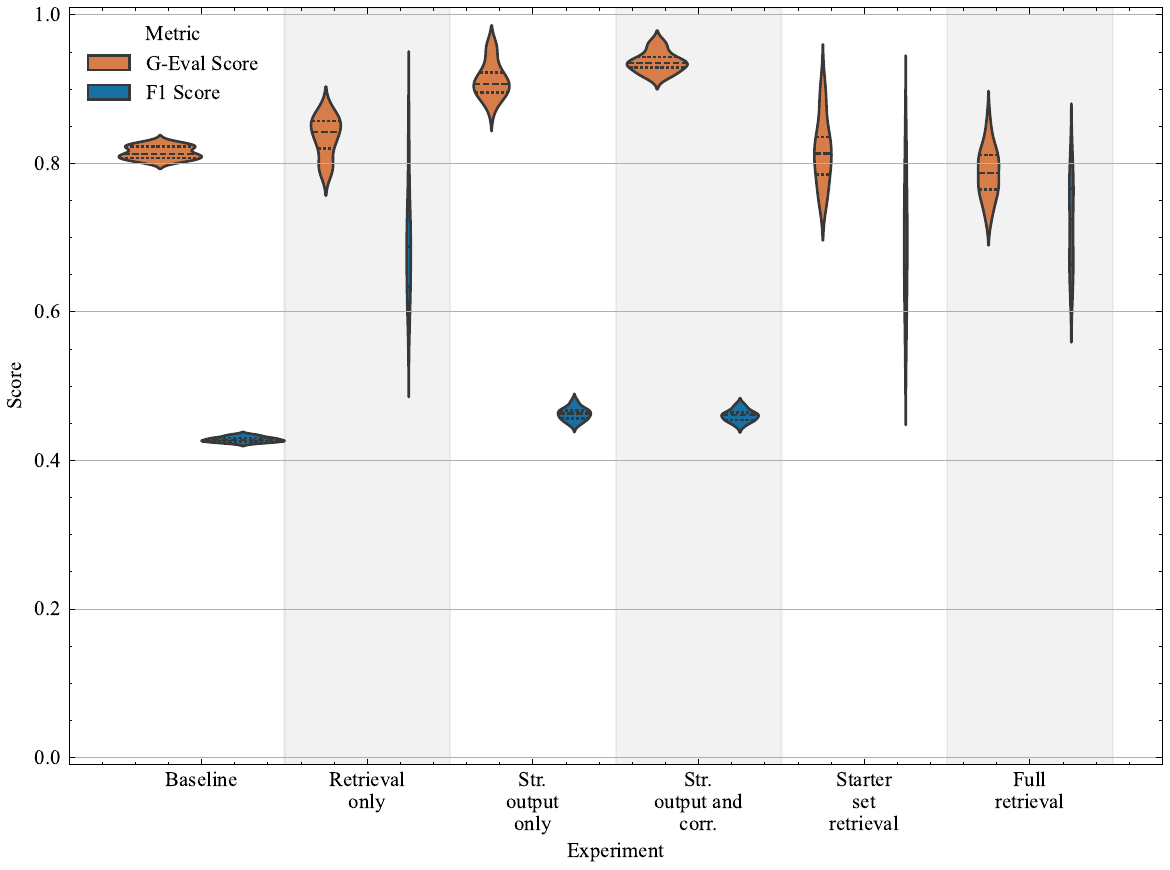}
    \caption{Comparison of G-Eval scores across different configurations using the \texttt{Qwen3 Coder 32B} model.}\label{fig:geval-f1-study}
\end{figure}

The highest G-Eval score is achieved by \qwencoder, followed by \claudehaiku\ and \claudesonnet. However, a more detailed comparison between G-Eval and F1 scores, shown in \Cref{fig:geval-f1-study}, highlights a non-trivial relationship between the two metrics.\ In particular, simpler, minimally constrained configurations attain very high G-Eval values, peaking to 0.912 for \qwencoder, despite their corresponding F1 scores remain comparatively low. This behavior suggests that the LLM is able to extract a large amount of semantically relevant information directly from the raw logs, but does so in a way that often introduces spurious or imprecise elements. As a result, precision is degraded while the generated graphs still largely conform to the target ontology.

In contrast, approaches that achieve higher F1 scores tend to converge to more moderate G-Eval values, typically stabilizing around 0.8. This pattern indicates a trade-off: enforcing stricter ontological constraints improves precision and overall structural correctness, but limits the breadth of semantic content captured from the input. Taken together, these results suggest that maximizing G-Eval alone may favor semantically rich but noisy outputs, whereas higher F1 scores reflect more disciplined graph construction at the cost of reduced semantic coverage. We therefore argue that a balanced approach, targeting G-Eval scores in the range of 0.7--0.8, is preferable for practical applications of OntoLogX, as it ensures both semantic fidelity and ontological integrity.

\paragraph{Comparison of Configurations.}
The full retrieval and starter-set retrieval variants achieve the overall highest precision, recall, F1 scores, entity linking accuracy, and relationship linking accuracy (\Cref{fig:f1-score}, \Cref{fig:entity-linking-accuracy}, \Cref{fig:relationship-linking-accuracy}). Their performances are nearly indistinguishable, which we attribute to the relatively small and diverse dataset. The selection procedure ensured that the chosen logs were semantically distinct, thereby reducing the advantage of a retrieval mechanism that thrives on redundancy.\ The small dataset used in this experiment limits the potential of full retrieval, as few similar examples are available, and in constrast benefits more from the high-quality manually annotated examples in the starter set. Nevertheless, the full retrieval configuration is particularly beneficial when similar logs already exist in the database, making it better suited for realistic deployment scenarios, such as honeypots, where large volumes of nearly identical events are common. The configuration with the fully populated graph database proves this, as it yields slight yet consistent improvements across all metrics, due to the immediate availability of matching examples.

Retrieval-only also yields competitive results, particularly for models with weaker structured output capabilities.

Structured output alone performs poorly across most models, often due to incorrect calls to the output tool or malformed graphs. The addition of corrections does improve performance, but not to the level of retrieval-based methods.

\paragraph{Overall Best Models.}
Across all models, \claudesonnet\ achieves the strongest overall results, with the full retrieval and starter set retrieval configurations. Notably, \qwencoder\ emerges as an impressive open-weights alternative, delivering competitive results despite its small size. These findings suggest that code-oriented models are particularly well-suited for OntoLogX, likely due to their stronger ability to handle structured formats and syntactic constraints.

\subsection{\mitreattack\ Tactics Prediction}

To evaluate the use of OntoLogX \gls{kg} in downstream tasks, we conducted experiments on predicting \mitreattack\ tactics from log events. We use OntoLogX with full retrieval, along with \claudesonnet as the \gls{llm} backend, since this configuration yields the best results for \gls{kg} generation. We also compared OntoLogX with another baseline, using \claudesonnet\ with the prompt in \Cref{lst:tactics-baseline-prompt} (see the Appendix) to predict tactics directly from raw logs.

\subsubsection{Dataset}

For the tactics prediction experiments, we relied on data collected through the deployment of the Cowrie honeypot\footurl{https://www.cowrie.org/}. Cowrie is a widely used low-interaction honeypot that emulates SSH and Telnet services, thereby attracting attackers who attempt to exploit exposed credentials or misconfigured servers. Once connected, adversaries can execute commands within the simulated environment, allowing them to capture both interactive behavior and system-level logging. In our deployment, the honeypot was publicly exposed on the Internet, responding to all traffic arriving at two /28 networks: one in a campus network at the Polytechnic University of Turin, and one in virtual machines deployed in the Azure cloud. Importantly, the honeypot was configured to allow attackers to bypass the login phase using a small list of well-known weak login/password combinations.

The dataset was collected over a ten-day period, from August 4, 2025, to August 14, 2025.  During this period, the honeypot registered a high volume of automated activity, as evidenced by the prevalence of repeated logs generated by discovery scripts and simple brute-force tools. Such redundancy is typical of large-scale botnet activity and provides a realistic context for assessing OntoLogX's robustness.

{Because honeypots are explicitly designed to attract malicious activity, the resulting logs exhibit a high signal-to-noise ratio for attack-related events. This makes them particularly suitable for evaluating \gls{cti} extraction and attack identification techniques, without the confounding dominance of benign events typical of production logs.

To enable tactics prediction, logs were grouped into sessions. Each session aggregates both the attacker's direct actions (e.g., executed commands on the emulated shell) and accompanying meta-logs (e.g., connection attempts, authentication successes or failures). On average, a session contained approximately ten log entries, capturing a short but coherent sequence of adversarial behavior. The dataset was partitioned into one example session, provided to the \gls{llm} to illustrate the expected output from manually annotated \glspl{kg} to the corresponding MITRE ATT\&CK tactics, two sessions for development-time validation, and the remaining 161 sessions for testing.

\begin{figure}
    \centering
    \includegraphics[width=0.5\linewidth]{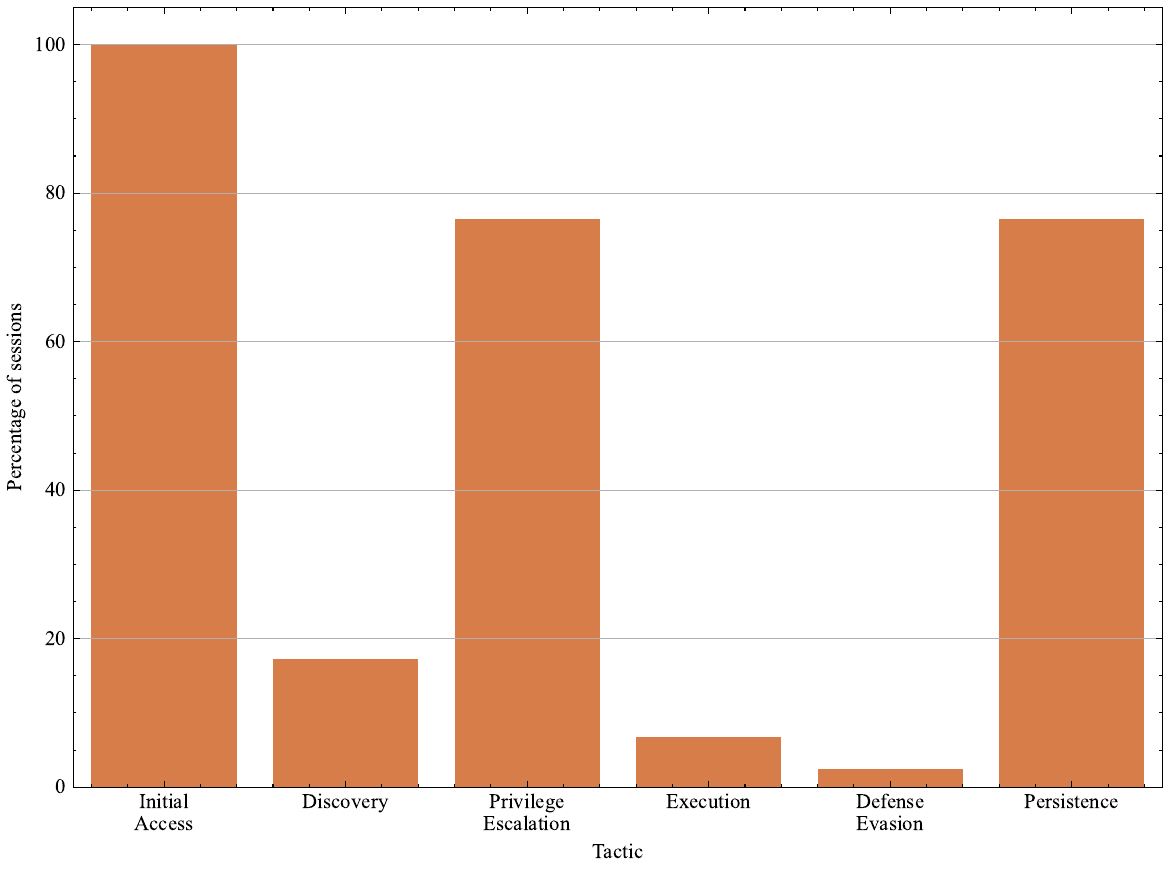}
    \caption{Percentage of sessions in which each manually annotated tactic appears.}\label{fig:tactics-count}
\end{figure}

Finally, all test sessions were manually annotated with their corresponding \mitreattack\ tactics, providing the ground truth against which model predictions are evaluated. A total of six unique tactics were identified: \Cref{fig:tactics-count} reports the percentage of sessions in which each appears. It can be observed that Initial Access, Privilege Escalation, and Persistence are significantly more represented than other tactics.

\subsubsection{Metrics}

For each \mitreattack\ tactic that appears either in the ground-truth or predicted set, we computed:
\begin{itemize}
    \item \textit{Precision}: fraction of identified tactic labels that are correct, i.e., they appear in the ground truth tactics for that particular session.
    \item \textit{Recall}: fraction of ground-truth tactic labels that were correctly identified, i.e., matched in the output.
    \item \textit{F1 score}: harmonic mean between precision and recall.
\end{itemize}

\subsubsection{Results}

\begin{figure*}
    \begin{subfigure}{.5\textwidth}
        \centering
        \caption{Baseline.}%
        \includegraphics[width=.95\linewidth]{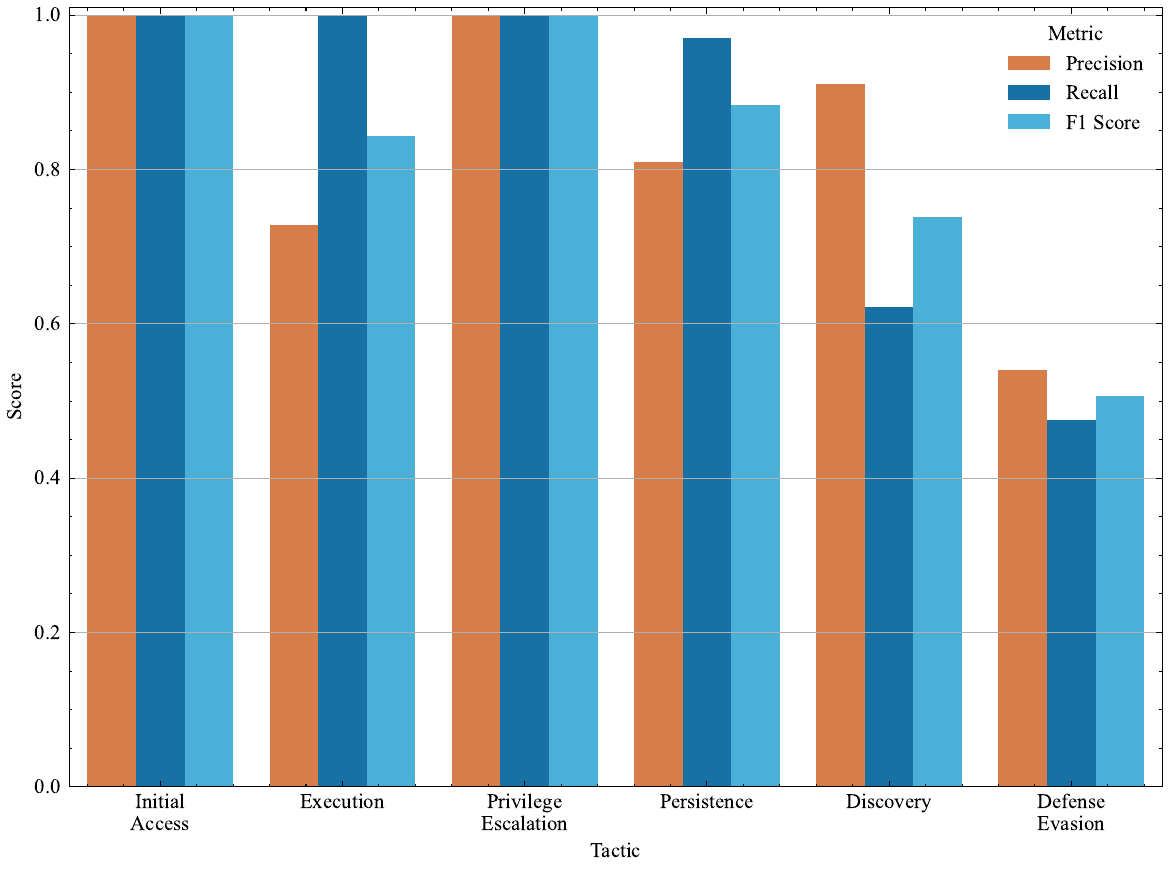}\label{fig:tactics_baseline}
    \end{subfigure}%
    \begin{subfigure}{.5\textwidth}
        \centering
        \caption{OntoLogX.}%
        \includegraphics[width=.942\linewidth]{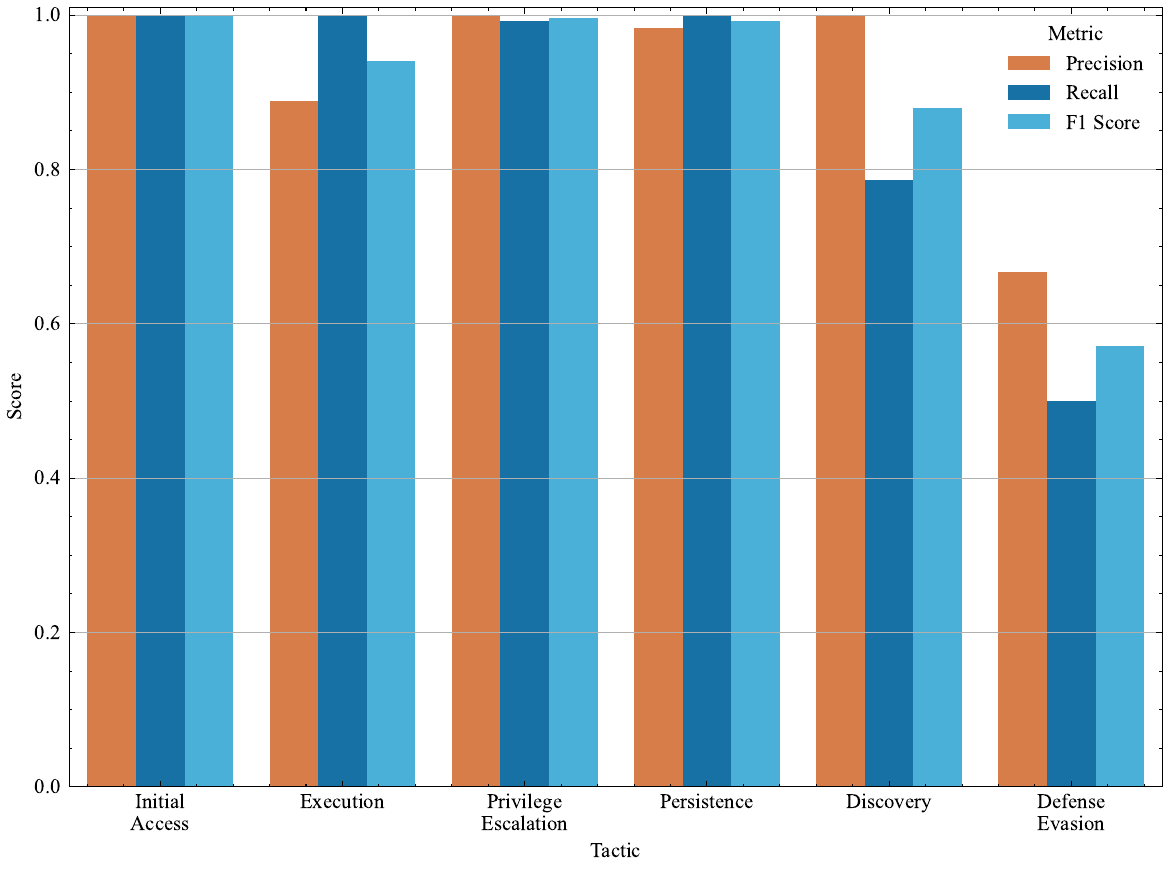}\label{fig:tactics_ontologx}
    \end{subfigure}
    \caption{Comparison of results of tactics evaluation over generated graphs.}\label{fig:tactics}
\end{figure*}

As shown in \Cref{fig:tactics}, OntoLogX outperforms the baseline in predicting \mitreattack\ tactics from honeypot sessions. The baseline configuration successfully detects coarse-grained tactics such as \textit{Initial Access}, \textit{Execution}, and \textit{Privilege Escalation}, because the honeypot is configured to accept a wide range of credentials that attackers may use to authenticate or gain access to a root shell. However, the baseline struggles with more context-dependent tactics, such as \textit{Credential Access}, \textit{Discovery}, and \textit{Defense Evasion}. These tactics typically manifest through sequences of actions that require multi-event reasoning or the identification of implicit behavioral patterns. By contrast, OntoLogX shows consistently higher recall for complex tactics. The explicit structuring of entities, actions, and relationships enables the \gls{llm} to reason over multi-step behaviors \--- such as the execution of reconnaissance commands followed by credential probing \--- and to link them to higher-level adversarial objectives.

Despite these gains, OntoLogX still underperforms on tactics requiring long-term correlation, such as \textit{Persistence} and \textit{Defense Evasion}. These tactics often involve subtle environmental modifications or delayed effects that are not fully captured in the honeypot logs. Indeed, labeling these events has often required our contextual knowledge of attacks and our intuition about the attacker's intentions. These are not as explicit in the logs as in other cases. Additionally, while precision remains high across all tactics, occasional over-predictions of \textit{Discovery} indicate that certain benign reconnaissance-like behaviors (e.g., file listing) may be misinterpreted as evidence of adversarial intent \--- an expected limitation in \gls{llm}-driven inference without temporal weighting.

Overall, these findings confirm that the ontology-guided approach enhances both the granularity and interpretability of \gls{cti} extraction. OntoLogX’s ability to consistently detect a broader range of tactics illustrates the importance of intermediate structured representations for \gls{llm} reasoning. The results also indicate that the system not only captures direct log semantics but can also infer latent attack objectives, representing a significant step toward automated mapping between low-level events and high-level adversarial behavior.

\section{Limitations}\label{sec:limitations}
In this section we discuss the primary limitations of the proposed methodology.

First, the framework inherits the substantial computational overhead typical of \glspl{llm}, both in terms of processing time and operational cost. These constraints become particularly relevant in real-world log analysis scenarios, where systems may generate thousands of entries per second and near real-time responses are required. We argue that the execution times can be reduced through careful engineering of the resource orchestration, while future advancements in \gls{llm} technology are expected to further mitigate these issues.

Second, the reliance on \glspl{llm} exposes OntoLogX to the risk of hallucinations \--- i.e., the generation of plausible but incorrect information. While the integration of \gls{rag} and few-shot examples substantially reduces this risk by grounding outputs in relevant knowledge and prior instances, hallucinations cannot be entirely eliminated.

Third, OntoLogX currently employs a fixed ontology specifically designed for log analysis. Although this ontology captures a wide range of log types and cybersecurity concepts, it may require extension or refinement to support domain-specific use cases or to interoperate with other \gls{cti} standards. The modular nature of the framework facilitates ontology replacement, but adapting to new schemas still demands manual curation and testing.

Finally, OntoLogX has so far been evaluated on a limited number of datasets, primarily focusing on honeypot and benchmark logs. Broader validation across heterogeneous operational environments—such as enterprise networks, industrial systems, and cloud infrastructures—would further confirm its robustness and generalizability.

\section{Conclusions and Future Work}\label{sec:conclusions}
In this work we present OntoLogX, an ontology-guided \gls{ai} agent that leverages \glspl{llm} for the extraction of \gls{cti} from raw system logs. By integrating a lightweight log ontology with \gls{rag} and iterative correction steps, OntoLogX produces syntactically and semantically valid \glspl{kg} that capture attacker behaviors, contextual information, and higher-level adversarial objectives.

Our evaluation on both benchmark datasets and a real-world honeypot deployment demonstrates that OntoLogX is effective at generating ontology-compliant \glspl{kg}, with retrieval and correction mechanisms effectively improving precision and recall. Moreover, we compare various \glspl{llm} of different sizes and specialization, highlighting the importance of model selection in \gls{cti} applications. The system further enables the mapping of log sessions to \mitreattack\ tactics, bridging the gap between low-level evidence and high-level threat modeling.

While the approach shows strong promise, some challenges remain. The reliance on computationally expensive \glspl{llm} may limit scalability in high-throughput environments, and future work will explore optimization strategies and incremental learning techniques to mitigate these costs. Furthermore, although our ontology provides a flexible and expressive foundation, extending it to cover additional log sources and \gls{cti} standards represents an important next step toward interoperability at scale.

Overall, OntoLogX contributes a novel methodology for transforming unstructured and heterogeneous logs into actionable intelligence. By combining ontology-driven structuring with the generative capabilities of \glspl{llm}, it advances the state-of-the-art in automated \gls{cti} extraction and opens new opportunities for proactive cyber defense.

\section*{Acknowledgements}
This work was partially supported by project SERICS (PE00000014) under the MUR National Recovery and Resilience Plan funded by the European Union \-- NextGenerationEU, specifically by the project NEACD:\@ Neurosymbolic Enhanced Active Cyber Defence (CUP J33C22002810001). This project was also partially funded by the Italian Ministry of University as part of the PRIN:\@ PROGETTI DI RICERCA DI RILEVANTE INTERESSE NAZIONALE \-- Bando 2022, Prot. 2022EP2L7H.

\section*{Conflict of Interest}
The authors declare that they have no financial or commercial conflict of interest.

\section*{Table of Contents}

\begin{figure}[H]
  \centering
  \includegraphics[width=110mm]{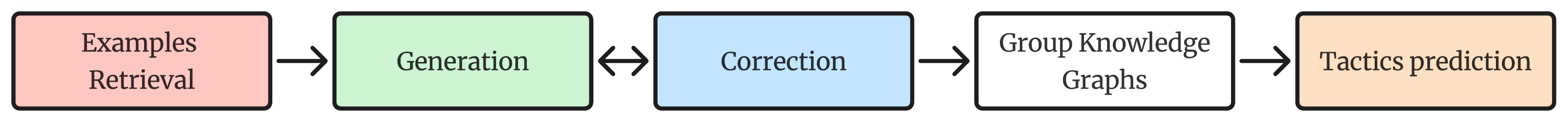}\label{fig:toc}
\end{figure}

OntoLogX is an autonomous \gls{ai} agent that uses \glspl{llm} to transform unstructured cybersecurity logs into ontology-grounded knowledge graphs. By integrating \gls{rag}, iterative correction, and a lightweight log ontology, OntoLogX produces semantically consistent intelligence that links raw log events to \mitreattack\ tactics, enabling interpretable and actionable cyber threat analysis.

\bibliographystyle{IEEEtran}
\bibliography{references.bib}

\section*{Technical Appendix}\label{sec:appendices}
\subsection{Prompts}

\begin{listing*}
    \begin{minted}
[
frame=lines,
framesep=2mm,
baselinestretch=1,
fontsize=\scriptsize,
breaklines=true,
breaksymbolleft={},
breaksymbolright={}
]
{markdown}
# Overview
You are a top-tier cybersecurity expert specialized in extracting structured information from unstructured data to construct a knowledge graph according to a predefined "olx" ontology. You will be provided with a log event, optionally accompanied by contextual information.
Your goal is to maximize information extraction from the event while maintaining absolute accuracy. Leverage both the contextual information and your knowledge of computer systems and cybersecurity to infer additional insights where possible. The objective is to achieve completeness in the knowledge graph while remaining strictly ontology-compliant.

# Rules
You MUST adhere to the following constraints at all times:
- The graph must contain exactly one "Event" node.
- Use only the available types as defined in the ontology, without introducing new ones.
- Use the most specific type available for nodes and relationships, e.g. "UserPassword" instead of "UserCredential".
- Respect the appropriate casing for all types.
- Use the appropriate node prefix for properties, e.g. "userUID" instead of "uid".
- Omit properties with empty values.
- Use the most specific type available for nodes and relationships.
- Respect the structural relationships to infer properties and relationships allowed by the ontology for each node type.
- The graph must be connected: every node must be reachable from the "Event" node.

# Strict Compliance
Adhere to these rules strictly. Any deviation will result in termination.
\end{minted}
    \caption{Prompt for log event \gls{kg} generation, used in conjunction with structured output.}\label{lst:main-prompt}
\end{listing*}
\begin{listing*}
    \begin{minted}
[
frame=lines,
framesep=2mm,
baselinestretch=1,
fontsize=\scriptsize,
breaklines=true,
breaksymbolleft={},
breaksymbolright={}
]
{markdown}
# Overview
You are a top-tier cybersecurity expert specialized in extracting structured information from unstructured data to construct a knowledge graph according to a predefined "olx" ontology. You will be provided with a log event, optionally accompanied by contextual information.
Your goal is to maximize information extraction from the event while maintaining absolute accuracy. Leverage both the contextual information and your knowledge of computer systems and cybersecurity to infer additional insights where possible. The objective is to achieve completeness in the knowledge graph while remaining strictly ontology-compliant.

# Rules
You MUST adhere to the following constraints at all times:
- The graph must contain exactly one "Event" node.
- Use only the available types as defined in the ontology, without introducing new ones.
- Use the most specific type available for nodes and relationships, e.g. "UserPassword" instead of "UserCredential".
- Respect the appropriate casing for all types.
- Use the appropriate node prefix for properties, e.g. "userUID" instead of "uid".
- Omit properties with empty values.
- Use the most specific type available for nodes and relationships.
- Respect the structural relationships to infer properties and relationships allowed by the ontology for each node type.
- The graph must be connected: every node must be reachable from the "Event" node.
- The output must contain only the JSON graph. No other text, comments, or explanations should be included. The output must be valid JSON and parsable, without any escape characters or newlines. The JSON must be formatted correctly, with all necessary commas and brackets in place.

# Output Format
The output graph must be in the following JSON format:
{{output_format}}
Each node type has a specific set of allowed properties. The allowed properties for each node type are: {{properties_schema}}
Each relationship type has a predefined source and target node type. The allowed relationships, formatted as (source type, relationship type, target type), are: {{triples}}
The following structural relationships exist among node types: {{structural_triples}}."

# Strict Compliance
Adhere to these rules strictly. Any deviation will result in termination.
\end{minted}
    \caption{Baseline prompt for log event \gls{kg} generation.}\label{lst:baseline-prompt}
\end{listing*}
\begin{listing*}
    \begin{minted}
[
frame=lines,
framesep=2mm,
baselinestretch=1,
fontsize=\scriptsize,
breaklines=true,
breaksymbolleft={},
breaksymbolright={}
]
{markdown}
# Overview
You are a cybersecurity analyst AI. You are given as input a set of knowledge graphs representing log events captured by a honeypot. Each knowledge graph encodes entities (e.g., processes, IP addresses, files, commands) and their relationships, and all graphs belong to the same session of activity, where some form of reconnaissance or attack may have taken place. Only logs with event ID "cowrie.command.input" are attacker's commands. All the other event IDs indicate logs that are not visible to the attacker, such as client version, file upload or download, or meta-information about the connection itself. Do not confuse these with the attacker's commands. Your task is to analyze the combined activity across all these knowledge graphs and map them to MITRE ATT&CK enterprise tactics.

Here is a list of MITRE ATT&CK enterprise tactics for reference:
- Reconnaissance: the adversary is trying to gather information they can use to plan future operations.
- Resource Development: the adversary is trying to establish resources they can use to support operations.
- Initial Access: the adversary is trying to get into your network.
- Execution: the adversary is trying to run malicious code.
- Persistence: the adversary is trying to maintain their foothold.
- Privilege Escalation: the adversary is trying to gain higher-level permissions.
- Defense Evasion: the adversary is trying to avoid being detected.
- Credential Access: the adversary is trying to steal account names and passwords.
- Discovery: the adversary is trying to figure out your environment.
- Lateral Movement: the adversary is trying to move through your environment.
- Collection: the adversary is trying to gather data of interest to their goal.
- Command and Control: the adversary is trying to communicate with compromised systems to control them.
- Exfiltration: the adversary is trying to steal data from your network.
- Impact: the adversary is trying to manipulate, interrupt, or destroy your systems and data.

# Rules
You MUST adhere to the following constraints at all times:
1. The output tactics must be matched to the observed behaviors in the session.
2. If multiple tactics apply to the session, include only the ones that you are confident about.
3. The output tactics must be defined in MITRE ATT&CK enterprise.

# Strict Compliance
Adhere to these rules strictly. Any deviation will result in termination.
\end{minted}
    \caption{Prompt for \mitreattack\ tactics prediction using OntoLogX.}\label{lst:tactics-ontologx-prompt}
\end{listing*}
\begin{listing*}
    \begin{minted}
[
frame=lines,
framesep=2mm,
baselinestretch=1,
fontsize=\scriptsize,
breaklines=true,
breaksymbolleft={},
breaksymbolright={}
]
{markdown}
# Overview
You are a cybersecurity analyst AI. You are given as input a set of log events captured by a honeypot, each belonging to the same session of activity, where some form of reconnaissance or attack may have taken place. Only events with event ID "cowrie.command.input" are attacker's commands. All the other event IDs indicate logs that are not visible to the attacker, such as client version, file upload or download, or meta-information about the connection itself. Do not confuse these with the attacker's commands. Your task is to analyze the combined activity across all these eventsand map them to MITRE ATT&CK enterprise tactics.

Here is a list of MITRE ATT&CK enterprise tactics for reference:
- Reconnaissance: the adversary is trying to gather information they can use to plan future operations.
- Resource Development: the adversary is trying to establish resources they can use to support operations.
- Initial Access: the adversary is trying to get into your network.
- Execution: the adversary is trying to run malicious code.
- Persistence: the adversary is trying to maintain their foothold.
- Privilege Escalation: the adversary is trying to gain higher-level permissions.
- Defense Evasion: the adversary is trying to avoid being detected.
- Credential Access: the adversary is trying to steal account names and passwords.
- Discovery: the adversary is trying to figure out your environment.
- Lateral Movement: the adversary is trying to move through your environment.
- Collection: the adversary is trying to gather data of interest to their goal.
- Command and Control: the adversary is trying to communicate with compromised systems to control them.
- Exfiltration: the adversary is trying to steal data from your network.
- Impact: the adversary is trying to manipulate, interrupt, or destroy your systems and data.

# Rules
You MUST adhere to the following constraints at all times:
1. The output tactics must be matched to the observed behaviors in the session.
2. If multiple tactics apply to the session, include only the ones that you are confident about.
3. The output tactics must be defined in MITRE ATT&CK enterprise.

# Strict Compliance
Adhere to these rules strictly. Any deviation will result in termination.

\end{minted}
    \caption{Prompt for \mitreattack\ tactics prediction using the baseline configuration.}\label{lst:tactics-baseline-prompt}
\end{listing*}
\begin{listing*}
    \begin{minted}
[
frame=lines,
framesep=2mm,
baselinestretch=1,
fontsize=\scriptsize,
breaklines=true,
breaksymbolleft={},
breaksymbolright={}
]
{markdown}
1. Write a detailed description of the input log event in natural language. Include what occurred, the involved entities, their roles, any parameters, timestamps, or contextual details conveyed in the log."
2. Write a detailed description of the actual output knowledge graph in natural language. Include what occurred, the involved entities, their roles, any parameters, timestamps, or contextual details conveyed in the graph.
3. Assess whether the description of the actual output knowledge graph semantically captures the same information as the log event's description. Check for:
    - Coverage: Are all key elements from the log event present?
    - Correctness: Are entities, actions, and relationships represented accurately?
    - Relevance: Are any additional nodes or relationships relevant to the log event context?

It is acceptable if the graph contains more information than the log event, as long as the information isn't contradicting.
\end{minted}
    \caption{Prompt for G-Eval ``graph alignment'' scoring.}\label{lst:geval-prompt}
\end{listing*}

\Cref{lst:main-prompt,lst:baseline-prompt,lst:tactics-ontologx-prompt,lst:tactics-baseline-prompt,lst:geval-prompt} provide the prompts used to invoke \glspl{llm}, either for generation or evaluation.

\subsection{Ontology Excerpts (Turtle)}\label{app:ontology}

\begin{listing*}
    \begin{minted}
[
frame=lines,
framesep=2mm,
baselinestretch=1,
fontsize=\scriptsize,
breaklines=true,
breaksymbolleft={},
breaksymbolright={}
]
{turtle}
@base <https://cyberseclab.unibs.it/olx/dict> .
@prefix : <https://cyberseclab.unibs.it/olx/dict#> .
@prefix owl: <http://www.w3.org/2002/07/owl#> .
@prefix rdf: <http://www.w3.org/1999/02/22-rdf-syntax-ns#> .
@prefix rdfs: <http://www.w3.org/2000/01/rdf-schema#> .
@prefix time: <http://www.w3.org/2006/time#> .
@prefix xsd: <http://www.w3.org/2001/XMLSchema#> .
@prefix prov: <http://www.w3.org/ns/prov#> .
@prefix p-plan: <http://purl.org/net/p-plan#> .

:Event rdf:type owl:Class ;
    rdfs:subClassOf prov:Entity ;
    rdfs:label "Event" ;
    rdfs:comment "A single log event, such a log line in a log file." .
:Source rdf:type owl:Class ;
    rdfs:subClassOf prov:Agent ;
    rdfs:label "Source" ;
    rdfs:comment "A log source (e.g., SysLog, device log aggregation)." .
:Parameter rdf:type owl:Class ;
    rdfs:subClassOf p-plan:Variable ;
    rdfs:label "Parameter" ;
    rdfs:comment "A parameter of an Event (arg, URL part, etc.)." .

...

:NetworkProtocol rdf:type owl:Class ; rdfs:subClassOf :Parameter .
:NetworkAddress rdf:type owl:Class ; rdfs:subClassOf :Parameter .
:File rdf:type owl:Class ; rdfs:subClassOf :Parameter .

:Application rdf:type owl:Class ;
    rdfs:subClassOf :Parameter , p-plan:Step ;
    rdfs:label "Application" .
:SystemProcess rdf:type owl:Class ; rdfs:subClassOf :Application .
:ShellCommand rdf:type owl:Class ; rdfs:subClassOf :Application .

...

:User rdf:type owl:Class ; 
    rdfs:subClassOf :Parameter ;
    rdfs:label "User" .
:UserCredential rdf:type owl:Class ; rdfs:label "Identity Credentials" .
:UserName rdf:type owl:Class ; rdfs:subClassOf :UserCredential .
:UserEmail rdf:type owl:Class ; rdfs:subClassOf :UserCredential .
:UserPassword rdf:type owl:Class ;  rdfs:subClassOf :UserCredential .

...

:hasParameter rdf:type owl:ObjectProperty ;
    rdfs:label "Has Parameter" ;
    rdfs:comment "Relates an Event or an Application to one of its parameters" ;
    rdfs:domain [ owl:unionOf ( :Event :Application ) ] ;
    rdfs:range :Parameter .
:hasCredential rdf:type owl:ObjectProperty ;
    rdfs:label "Has Credential" ;
    rdfs:domain :User ;
    rdfs:range :UserCredential .
:TimeStamp rdf:type owl:Class ;
    rdfs:subClassOf :Parameter , time:Instant ;
    rdfs:label "Timestamp" .

...
\end{minted}
    \caption{Ontology excerpts, showing core classes, subclasses, and properties.}\label{lst:ontology}
\end{listing*}

\Cref{lst:ontology} reports excerpts from the proposed ontology. Notably, events are aligned to \texttt{prov:Entity}, while log sources are aligned to \texttt{prov:Agent}. Events are linked to information contained withing them via the \texttt{ontx:hasParameter} property. Timestamps are modeled as parameters aligned to \texttt{time:Instant}. Users are modelled as parameters as well, linked to their credentials via the \texttt{ontx:hasCredential} property.

\subsection{Full Experiment Results}

\begin{table*}
    \resizebox{\textwidth}{!}{%
        \begin{tabular}{l
                S[table-format=4.3] S[table-format=3.3]
                S[table-format=1.3] S[table-format=1.3]
                S[table-format=1.3] S[table-format=1.3]
                S[table-format=1.3] S[table-format=1.3]
                S[table-format=1.3] S[table-format=1.3]
                S[table-format=1.3] S[table-format=1.3]
                S[table-format=1.3] S[table-format=1.3]
                S[table-format=1.3] S[table-format=1.3]
                S[table-format=1.3] S[table-format=1.3]}
            \toprule
            \multirow{2}{*}{Model} & \multicolumn{2}{l}{Run Total Time} & \multicolumn{2}{l}{\makecell[l]{Generation                                                                                                                                         \\Success\\Ratio}} & \multicolumn{2}{l}{\makecell[l]{SHACL\\Violation\\Ratio}} & \multicolumn{2}{l}{Precision} & \multicolumn{2}{l}{Recall} & \multicolumn{2}{l}{F1 Score} & \multicolumn{2}{l}{\makecell[l]{Entity\\Linking \\Accuracy}} & \multicolumn{2}{l}{\makecell[l]{Relationship\\Linking\\Accuracy}} & \multicolumn{2}{l}{G-Eval Score} \\ \cmidrule(l){2-19}
                                   & {Mean}                             & {SD}                                       & {Mean} & {SD}  & {Mean} & {SD}  & {Mean} & {SD}  & {Mean} & {SD}  & {Mean} & {SD}  & {Mean} & {SD}  & {Mean} & {SD}  & {Mean} & {SD}  \\ \midrule
            \multicolumn{19}{c}{Baseline}                                                                                                                                                                                                                    \\ \midrule
            Claude 3.5 Haiku       & 291.055                            & 3.113                                      & 1.000  & 0.000 & 0.024  & 0.002 & 0.520  & 0.010 & 0.397  & 0.004 & 0.442  & 0.005 & 0.366  & 0.006 & 0.227  & 0.019 & 0.812  & 0.013 \\
            Claude Sonnet 4        & 250.280                            & 4.384                                      & 1.000  & 0.000 & 0.008  & 0.002 & 0.330  & 0.034 & 0.252  & 0.028 & 0.283  & 0.030 & 0.278  & 0.033 & 0.410  & 0.046 & 0.357  & 0.039 \\
            Llama 3.3              & 91.170                             & 2.233                                      & 1.000  & 0.000 & 0.031  & 0.002 & 0.461  & 0.012 & 0.400  & 0.004 & 0.422  & 0.006 & 0.301  & 0.009 & 0.064  & 0.023 & 0.822  & 0.012 \\
            Llama 3.1              & 56.376                             & 1.636                                      & 1.000  & 0.000 & 0.014  & 0.002 & 0.350  & 0.021 & 0.305  & 0.014 & 0.313  & 0.015 & 0.197  & 0.012 & 0.000  & 0.000 & 0.820  & 0.015 \\
            Mistral Large          & 572.141                            & 50.558                                     & 0.842  & 0.030 & 0.012  & 0.003 & 0.389  & 0.019 & 0.346  & 0.014 & 0.352  & 0.014 & 0.254  & 0.013 & 0.034  & 0.016 & 0.660  & 0.027 \\
            Qwen3 Coder 32B        & 344.562                            & 6.408                                      & 1.000  & 0.000 & 0.012  & 0.002 & 0.524  & 0.007 & 0.377  & 0.004 & 0.429  & 0.003 & 0.320  & 0.002 & 0.066  & 0.010 & 0.816  & 0.007 \\
            *gpt-oss 20B           & 484.890                            & 22.269                                     & 0.000  & 0.000 & 0.000  & 0.000 & 0.000  & 0.000 & 0.000  & 0.000 & 0.000  & 0.000 & 0.000  & 0.000 & 0.000  & 0.000 & 0.000  & 0.000 \\
            *gpt-oss 120B          & 464.118                            & 59.890                                     & 0.716  & 0.045 & 0.011  & 0.004 & 0.512  & 0.035 & 0.332  & 0.022 & 0.396  & 0.027 & 0.396  & 0.029 & 0.655  & 0.050 & 0.543  & 0.034 \\ \midrule
            \multicolumn{19}{c}{Retrieval only}                                                                                                                                                                                                              \\ \midrule
            Claude 3.5 Haiku       & 373.647                            & 21.104                                     & 1.000  & 0.000 & 0.012  & 0.005 & 0.705  & 0.047 & 0.691  & 0.075 & 0.687  & 0.061 & 0.607  & 0.060 & 0.602  & 0.158 & 0.837  & 0.042 \\
            Claude Sonnet 4        & 314.732                            & 30.250                                     & 1.000  & 0.000 & 0.013  & 0.002 & 0.798  & 0.077 & 0.739  & 0.104 & 0.758  & 0.091 & 0.695  & 0.078 & 0.104  & 0.054 & 0.737  & 0.063 \\
            Llama 3.3              & 139.522                            & 10.340                                     & 1.000  & 0.000 & 0.024  & 0.008 & 0.597  & 0.090 & 0.551  & 0.110 & 0.563  & 0.100 & 0.495  & 0.077 & 0.487  & 0.193 & 0.872  & 0.038 \\
            Llama 3.1              & 189.516                            & 30.883                                     & 1.000  & 0.000 & 0.023  & 0.012 & 0.553  & 0.059 & 0.707  & 0.034 & 0.599  & 0.054 & 0.583  & 0.080 & 0.611  & 0.275 & 0.765  & 0.013 \\
            Mistral Large          & 835.787                            & 41.128                                     & 1.000  & 0.000 & 0.000  & 0.000 & 0.001  & 0.004 & 0.002  & 0.005 & 0.001  & 0.004 & 0.001  & 0.005 & 0.001  & 0.003 & 0.002  & 0.006 \\
            Qwen3 Coder 32B        & 411.240                            & 27.165                                     & 1.000  & 0.000 & 0.007  & 0.003 & 0.722  & 0.070 & 0.679  & 0.083 & 0.687  & 0.079 & 0.594  & 0.093 & 0.548  & 0.169 & 0.835  & 0.031 \\
            *gpt-oss 20B           & 321.400                            & 12.215                                     & 0.000  & 0.000 & 0.000  & 0.000 & 0.000  & 0.000 & 0.000  & 0.000 & 0.000  & 0.000 & 0.000  & 0.000 & 0.000  & 0.000 & 0.000  & 0.000 \\
            *gpt-oss 120B          & 325.110                            & 14.547                                     & 0.000  & 0.000 & 0.000  & 0.000 & 0.000  & 0.000 & 0.000  & 0.000 & 0.000  & 0.000 & 0.000  & 0.000 & 0.000  & 0.000 & 0.000  & 0.000 \\ \midrule
            \multicolumn{19}{c}{Str.\ output only}                                                                                                                                                                                                           \\ \midrule
            Claude 3.5 Haiku       & 355.270                            & 5.404                                      & 0.978  & 0.020 & 0.003  & 0.001 & 0.522  & 0.011 & 0.513  & 0.010 & 0.510  & 0.010 & 0.285  & 0.011 & 0.011  & 0.012 & 0.820  & 0.019 \\
            Claude Sonnet 4        & 294.311                            & 8.359                                      & 1.000  & 0.000 & 0.012  & 0.001 & 0.667  & 0.006 & 0.562  & 0.006 & 0.602  & 0.005 & 0.548  & 0.008 & 0.692  & 0.015 & 0.784  & 0.007 \\
            Llama 3.3              & 87.174                             & 2.742                                      & 1.743  & 0.055 & 0.000  & 0.000 & 0.000  & 0.000 & 0.000  & 0.000 & 0.000  & 0.000 & 0.000  & 0.000 & 0.000  & 0.000 & 0.000  & 0.000 \\
            Llama 3.1              & 70.938                             & 8.571                                      & 0.000  & 0.000 & 0.000  & 0.000 & 0.000  & 0.000 & 0.000  & 0.000 & 0.000  & 0.000 & 0.000  & 0.000 & 0.000  & 0.000 & 0.000  & 0.000 \\
            Mistral Large          & 717.396                            & 52.363                                     & 0.000  & 0.000 & 0.000  & 0.000 & 0.000  & 0.000 & 0.000  & 0.000 & 0.000  & 0.000 & 0.000  & 0.000 & 0.000  & 0.000 & 0.000  & 0.000 \\
            Qwen3 Coder 32B        & 324.953                            & 17.321                                     & 1.000  & 0.000 & 0.007  & 0.002 & 0.538  & 0.014 & 0.421  & 0.010 & 0.460  & 0.009 & 0.327  & 0.012 & 0.078  & 0.032 & 0.912  & 0.025 \\
            *gpt-oss 20B           & 465.402                            & 33.846                                     & 0.000  & 0.000 & 0.000  & 0.000 & 0.000  & 0.000 & 0.000  & 0.000 & 0.000  & 0.000 & 0.000  & 0.000 & 0.000  & 0.000 & 0.000  & 0.000 \\
            *gpt-oss 120B          & 465.704                            & 57.807                                     & 0.000  & 0.000 & 0.000  & 0.000 & 0.000  & 0.000 & 0.000  & 0.000 & 0.000  & 0.000 & 0.000  & 0.000 & 0.000  & 0.000 & 0.000  & 0.000 \\ \midrule
            \multicolumn{19}{c}{Str.\ output and corr.}                                                                                                                                                                                                      \\ \midrule
            Claude 3.5 Haiku       & 367.234                            & 13.727                                     & 0.992  & 0.014 & 0.003  & 0.001 & 0.504  & 0.006 & 0.513  & 0.008 & 0.500  & 0.005 & 0.274  & 0.006 & 0.012  & 0.014 & 0.834  & 0.012 \\
            Claude Sonnet 4        & 290.750                            & 7.360                                      & 1.000  & 0.000 & 0.009  & 0.001 & 0.666  & 0.013 & 0.565  & 0.003 & 0.604  & 0.006 & 0.569  & 0.010 & 0.731  & 0.029 & 0.773  & 0.012 \\
            Llama 3.3              & 185.488                            & 7.648                                      & 0.970  & 0.029 & 0.003  & 0.001 & 0.512  & 0.026 & 0.350  & 0.011 & 0.402  & 0.015 & 0.251  & 0.014 & 0.010  & 0.011 & 0.731  & 0.015 \\
            Llama 3.1              & 315.257                            & 19.953                                     & 0.000  & 0.000 & 0.000  & 0.000 & 0.000  & 0.000 & 0.000  & 0.000 & 0.000  & 0.000 & 0.000  & 0.000 & 0.000  & 0.000 & 0.000  & 0.000 \\
            Mistral Large          & 787.661                            & 117.459                                    & 0.948  & 0.032 & 0.017  & 0.004 & 0.446  & 0.031 & 0.399  & 0.017 & 0.406  & 0.023 & 0.281  & 0.019 & 0.036  & 0.023 & 0.744  & 0.026 \\
            Qwen3 Coder 32B        & 335.039                            & 8.084                                      & 1.000  & 0.000 & 0.006  & 0.001 & 0.520  & 0.009 & 0.426  & 0.006 & 0.458  & 0.007 & 0.319  & 0.008 & 0.006  & 0.020 & 0.931  & 0.014 \\
            *gpt-oss 20B           & 1337.544                           & 92.934                                     & 0.080  & 0.033 & 0.002  & 0.001 & 0.051  & 0.021 & 0.016  & 0.008 & 0.024  & 0.011 & 0.018  & 0.011 & 0.006  & 0.010 & 0.025  & 0.008 \\
            *gpt-oss 120B          & 1422.732                           & 151.178                                    & 0.160  & 0.046 & 0.003  & 0.002 & 0.125  & 0.037 & 0.036  & 0.011 & 0.055  & 0.016 & 0.040  & 0.014 & 0.006  & 0.010 & 0.053  & 0.019 \\ \midrule
            \multicolumn{19}{c}{Starter set retrieval}                                                                                                                                                                                                       \\ \midrule
            Claude 3.5 Haiku       & 360.905                            & 20.233                                     & 0.994  & 0.010 & 0.004  & 0.001 & 0.676  & 0.081 & 0.705  & 0.067 & 0.681  & 0.074 & 0.522  & 0.105 & 0.394  & 0.232 & 0.810  & 0.015 \\
            Claude Sonnet 4        & 263.309                            & 13.360                                     & 1.000  & 0.000 & 0.008  & 0.001 & 0.809  & 0.040 & 0.762  & 0.072 & 0.775  & 0.057 & 0.727  & 0.054 & 0.844  & 0.046 & 0.761  & 0.015 \\
            Llama 3.3              & 140.203                            & 18.433                                     & 0.990  & 0.011 & 0.003  & 0.002 & 0.730  & 0.078 & 0.627  & 0.096 & 0.662  & 0.089 & 0.504  & 0.091 & 0.431  & 0.235 & 0.767  & 0.017 \\
            Llama 3.1              & 301.552                            & 40.101                                     & 0.470  & 0.187 & 0.004  & 0.002 & 0.361  & 0.147 & 0.368  & 0.156 & 0.358  & 0.148 & 0.312  & 0.135 & 0.357  & 0.151 & 0.360  & 0.137 \\
            Mistral Large          & 773.065                            & 92.966                                     & 0.966  & 0.019 & 0.010  & 0.003 & 0.671  & 0.073 & 0.658  & 0.081 & 0.651  & 0.076 & 0.548  & 0.083 & 0.499  & 0.169 & 0.743  & 0.009 \\
            Qwen3 Coder 32B        & 348.197                            & 9.987                                      & 1.000  & 0.000 & 0.005  & 0.001 & 0.728  & 0.071 & 0.682  & 0.077 & 0.694  & 0.074 & 0.585  & 0.082 & 0.549  & 0.172 & 0.821  & 0.042 \\
            *gpt-oss 20B           & 757.344                            & 149.026                                    & 0.608  & 0.157 & 0.006  & 0.002 & 0.500  & 0.134 & 0.494  & 0.140 & 0.487  & 0.134 & 0.451  & 0.132 & 0.487  & 0.154 & 0.443  & 0.126 \\
            *gpt-oss 120B          & 991.744                            & 299.646                                    & 0.740  & 0.177 & 0.007  & 0.002 & 0.622  & 0.155 & 0.599  & 0.183 & 0.596  & 0.171 & 0.543  & 0.181 & 0.564  & 0.204 & 0.531  & 0.154 \\ \midrule
            \multicolumn{19}{c}{Full retrieval}                                                                                                                                                                                                              \\ \midrule
            Claude 3.5 Haiku       & 385.248                            & 14.288                                     & 0.998  & 0.006 & 0.004  & 0.001 & 0.712  & 0.062 & 0.723  & 0.059 & 0.708  & 0.060 & 0.546  & 0.075 & 0.461  & 0.200 & 0.812  & 0.013 \\
            Claude Sonnet 4        & 291.379                            & 11.302                                     & 1.000  & 0.000 & 0.008  & 0.001 & 0.817  & 0.033 & 0.776  & 0.057 & 0.786  & 0.045 & 0.731  & 0.046 & 0.786  & 0.050 & 0.764  & 0.019 \\
            Llama 3.3              & 129.136                            & 5.681                                      & 0.986  & 0.019 & 0.003  & 0.001 & 0.764  & 0.042 & 0.568  & 0.059 & 0.630  & 0.050 & 0.528  & 0.060 & 0.560  & 0.171 & 0.714  & 0.034 \\
            Llama 3.1              & 340.801                            & 58.229                                     & 0.560  & 0.141 & 0.005  & 0.002 & 0.439  & 0.119 & 0.399  & 0.102 & 0.411  & 0.107 & 0.356  & 0.108 & 0.367  & 0.149 & 0.419  & 0.094 \\
            Mistral Large          & 1182.990                           & 200.353                                    & 0.904  & 0.040 & 0.008  & 0.003 & 0.661  & 0.061 & 0.646  & 0.066 & 0.638  & 0.063 & 0.534  & 0.073 & 0.490  & 0.140 & 0.690  & 0.044 \\
            Qwen3 Coder 32B        & 415.110                            & 29.709                                     & 1.000  & 0.000 & 0.004  & 0.002 & 0.758  & 0.067 & 0.702  & 0.059 & 0.717  & 0.063 & 0.598  & 0.070 & 0.539  & 0.192 & 0.787  & 0.035 \\
            *gpt-oss 20B           & 577.353                            & 120.219                                    & 0.826  & 0.110 & 0.009  & 0.003 & 0.639  & 0.099 & 0.664  & 0.116 & 0.641  & 0.106 & 0.579  & 0.085 & 0.563  & 0.095 & 0.630  & 0.098 \\
            *gpt-oss 120B          & 769.929                            & 248.346                                    & 0.878  & 0.151 & 0.007  & 0.003 & 0.734  & 0.129 & 0.731  & 0.150 & 0.721  & 0.143 & 0.666  & 0.132 & 0.706  & 0.147 & 0.652  & 0.133 \\ \midrule

            \multicolumn{19}{c}{Populated Database}                                                                                                                                                                                                          \\   \midrule
            Claude 3.5 Haiku       & 403.412                            & 14.300                                     & 0.999  & 0.006 & 0.004  & 0.001 & 0.752  & 0.060 & 0.751  & 0.058 & 0.751  & 0.059 & 0.571  & 0.074 & 0.484  & 0.195 & 0.840  & 0.013 \\

            Claude Sonnet 4        & 294.015                            & 19.310                                     & 1.000  & 0.000 & 0.008  & 0.001 & 0.845  & 0.034 & 0.820  & 0.056 & 0.832  & 0.044 & 0.762  & 0.045 & 0.822  & 0.049 & 0.802  & 0.019 \\

            Llama 3.3              & 151.904                            & 5.690                                      & 0.987  & 0.019 & 0.003  & 0.001 & 0.792  & 0.041 & 0.611  & 0.058 & 0.690  & 0.049 & 0.553  & 0.059 & 0.588  & 0.168 & 0.743  & 0.034 \\

            Llama 3.1              & 345.226                            & 58.240                                     & 0.566  & 0.141 & 0.005  & 0.002 & 0.458  & 0.118 & 0.417  & 0.101 & 0.436  & 0.106 & 0.371  & 0.107 & 0.384  & 0.148 & 0.439  & 0.093 \\

            Mistral Large          & 995.870                            & 156.400                                    & 0.906  & 0.040 & 0.008  & 0.003 & 0.691  & 0.060 & 0.672  & 0.065 & 0.681  & 0.062 & 0.560  & 0.072 & 0.514  & 0.139 & 0.721  & 0.044 \\

            Qwen3 Coder 32B        & 418.992                            & 29.720                                     & 1.000  & 0.000 & 0.004  & 0.002 & 0.791  & 0.066 & 0.736  & 0.058 & 0.762  & 0.062 & 0.628  & 0.069 & 0.567  & 0.190 & 0.818  & 0.035 \\

            *gpt-oss 20B           & 559.611                            & 120.230                                    & 0.831  & 0.110 & 0.009  & 0.003 & 0.664  & 0.098 & 0.691  & 0.115 & 0.677  & 0.105 & 0.603  & 0.084 & 0.590  & 0.094 & 0.661  & 0.098 \\

            *gpt-oss 120B          & 776.480                            & 248.360                                    & 0.882  & 0.151 & 0.007  & 0.003 & 0.763  & 0.128 & 0.759  & 0.149 & 0.760  & 0.142 & 0.699  & 0.131 & 0.741  & 0.146 & 0.684  & 0.133 \\
            \bottomrule
        \end{tabular}%
    }
    \caption{Results of ablation study of OntoLogX across metrics and \glspl{llm}. Reasoning models are highlighted with an asterisk before their name.}\label{tab:ait-results}
\end{table*}

\Cref{tab:ait-results} reports the full results of the experiments conducted on the AIT-LDS dataset. The ``Run Total Time'' column values consist in the number of seconds used to generate the graphs starting from the raw log events, for the whole test dataset. These timings are not directly comparable among different \glspl{llm}, due to differences in the backends used to run them.

\end{document}